\documentclass{article}

\usepackage[preprint]{neurips_2025}

\usepackage[utf8]{inputenc} 
\usepackage[T1]{fontenc}    
\usepackage{hyperref}       
\usepackage{url}            
\usepackage{booktabs}       
\usepackage{amsfonts}       
\usepackage{nicefrac}       
\usepackage{microtype}      
\usepackage[table]{xcolor}         
\usepackage{todonotes}
\usepackage{wrapfig}
\usepackage{listings}
\title{Randomly Sampled Language Reasoning Problems Elucidate Limitations of In-Context Learning}

\author{Kavi Gupta \\
Department of Electrical Engineering and Computer Science\\
Massachusetts Institute of Technology\\
Cambridge, MA, 02139, USA \\
\texttt{kavig@mit.edu} \\
\And
Kate Sanders \\
Department of Computer Science\\
Johns Hopkins University\\
Baltimore, MD 21218, USA\\
\texttt{ksande25@jhu.edu} \\
\AND
Armando Solar-Lezama \\
Department of Electrical Engineering and Computer Science\\
Massachusetts Institute of Technology\\
Cambridge, MA, 02139, USA \\
\texttt{asolar@csail.mit.edu} \\
}

\begin{document}

\maketitle

\begin{abstract}
While LLMs have revolutionized the field of machine learning due to their high performance on a strikingly wide range of problems, they are also known to hallucinate false answers and underperform on less canonical versions of the same tasks. There are several emerging theories of LLM performance, among them that LLMs lack world modeling ability, that they have an undesirable bias towards an autoregressive prior, and that they struggle on more novel problems. The existing literature on LLM input novelty has focused on tasks of relatively high complexity, studying perturbations of canonical but complex problems. In this paper, we attempt to minimize complexity in order to isolate novelty as a factor in LLM underperformance and investigate the power of in-context-learning. To this end, we consider an extremely simple domain: next token prediction on simple language tasks. The twist is that these language tasks are wholly unseen, as they are randomly drawn from a large, parsimoniously defined set of languages arising from simple grammar rules. This experimental setup allows us to evaluate ICL independently of models' parametric knowledge. We find that LLMs uniformly underperform n-gram models on this task, both when used as next token predictors and in chain-of-thought.
\end{abstract}
\section{Introduction}

\begin{wrapfigure}{R}{0.5\textwidth}
\includegraphics[width=\linewidth]{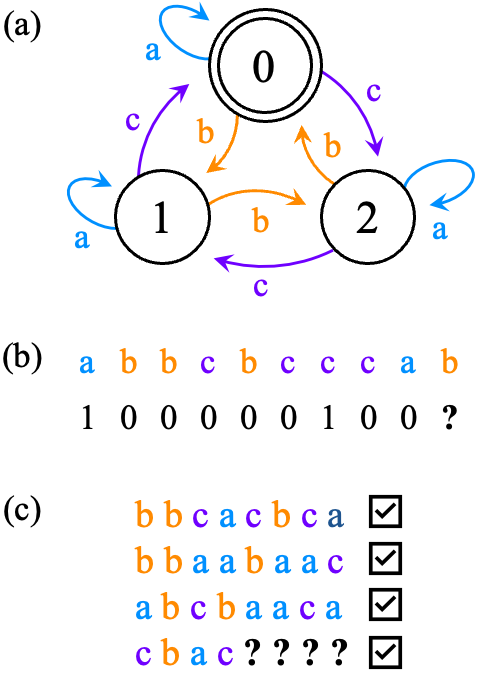}
\caption{We sample randomly generated languages to test LLMs by sampling deterministic finite automata (DFAs). (a) The DFA shown here, modeling the sum modulo 3 operation (with \texttt{abc} representing 0, 1, and 2 respectively), can be used to accept or reject strings from a 3-character alphabet. Accepted strings belong to the grammar; rejected strings do not. We evaluate models on their ability to (b) act as a transducer, recognizing strings that belong to the grammar, and (c) generate new strings following the grammar.}
\label{fig:intro-example}
\vspace{-5mm}
\end{wrapfigure}



One of the surprising capabilities of contemporary LLMs is their ability to perform in-context-learning (ICL), in which they learn by demonstration from provided input/output examplars to produce an appropriate output for a new but similarly structured input. This capability allows LLMs to generalize to tasks beyond interpolations of their training corpus, resulting in remarkably strong generalization capabilities. It is challenging to isolate the effects of ICL, as, given the large datasets LLMs are trained on, it is difficult to distinguish a model detecting novel patterns in examples from it being guided towards knowledge it gained during training. \citet{akyurek2022learning} demonstrate that, in specially trained transformers, ICL is ``true learning''. However, it remains unclear to what degree the performance of foundation-model LLMs can be attributed to strong ICL capabilities.

The task best suited to isolating LLM ICL performance should have the following properties: (1) It should be a language completion task within the expressive power of an LLM. LLMs are capable of many tasks, but they are primarily models of language, and as such, language tasks are the most fair evaluations. (2) It should not require sophisticated world modeling to solve. This helps us eliminate a possible source of underperformance distinct from ICL capability. (3) It should be selected at random in an unbiased manner to reduce the effect of bias from the training corpus.


To satisfy these properties, we propose the following general approach. First we define a large, exhaustive, and parsimoniously-defined space of languages that represents all languages of a certain difficulty level. Then, we sample random languages from this space. By sampling randomly, we can guarantee no bias towards canonical languages that might share structure with common ones in the training dataset. In this work, we use languages recognized by 3-state DFAs as these are the lowest nontrivial difficulty level. \footnote{This technique can be generalized to produce benchmarks of any difficulty level. For larger numbers of states, we would be able to guarantee that the majority of the exponentially large number of corresponding languages do not lie in the training dataset by a pigeonhole argument; unfortunately this does not apply to the relatively small set of 3-state DFAs (there are only 78786). However, they still represent an set of tasks of a particular difficulty level not biased towards the canonical.} Finally, to ensure we are not measuring world modeling performance, we compare to $n$-\textsc{Gram} baselines that are not capable of anything other than matching clusters of tokens.


Our results demonstrate that even for very simple language induction tasks that don't rely on world modeling or background knowledge, LLM ICL still underperforms simple language models when dealing with randomly sampled and likely unfamiliar problem instances. These results suggest that while LLMs can pick up some learning signal from examples in prompts, this in-context learning is not competitive with even very primitive forms of learning, and suggests that LLMs do not posses the ability to generalize to entirely novel language reasoning tasks.

In summary, we make the following contributions:
\begin{enumerate}
\item We introduce a benchmark for LLM ICL language reasoning evaluation, consisting of novel tasks.
\item We evaluate a suite of popular LLMs on instances of this benchmark and demonstrate that LLMs underperform compared to simple language model baselines.
\item We analyze the differences in behavior between these models, illustrating the influence of RLHF and chain-of-thought prompting on language reasoning capacity.
\end{enumerate}
\section{Related Work}

LLMs are known to fail in many cases, with some suggesting that these failures are due to lack of a world model \citep{valmeekam2022large} or ``embers'' of autoregression polluting non-autoregressive task performance \citep{mccoy2023embers}. Another theory is that of task novelty; that is, LLMs perform worse on tasks more dissimilar from those seen during training.

\subsection{Language Understanding and LLMs}
LLMs can be quite adept at generating programs in general-purpose programming languages~\citep{xu2022systematic}. In contrast, adapting models to understand domain-specific languages~\citep{mernik2005and} introduces unique problems such as navigating idiosyncratic syntax and semantics and leveraging sparse sample language data. To address these challenges, researchers have considered how well general-purpose LLMs can use language reasoning skills to quickly understand rare or unseen DSLs with only a small set of exemplars~\citep{joel2024survey}. While most work in this vein focuses on semantic parsing for downstream applications~\citep{lin2023few}, selecting exemplars~\citep{zhao2021calibrate}, and improving DSL recognition by leveraging more common languages~\citep{bogin2023leveraging}, experiments show strong baseline performance for LLM DSL recognition and parsing out-of-the-box~\citep{wang2024grammar}. Some have suggested that indicate that LLMs may possess emergent language reasoning abilities~\citep{milliere2024language}. 

Related lines of work are compositional generalization~\citep{xu2022compositional}, which assesses models' ability to organize known units into novel structures, and structural generalization~\citep{yao2022structural}, which assesses models' ability to recognize new structures. \citet{yao2022structural} show that smaller language models like BART and T5 can struggle on these tasks, but to our knowledge there are not comprehensive experiments extending this line of work to LLMs.

\subsection{Reasoning with LLMs}
Reasoning is one of many ``emergent abilities''~\citep{wei2022emergent} possibly possessed by LLMs~\citep{huang2022towards}, although the nonlinear dependence of such emergent abilities on model size is disputed~\citep{schaeffer2024emergent}. The chain-of-thought prompting technique~\citep{wei2022chain} has inspired a number of approaches to encourage the latent reasoning ability of models~\citep{yao2023tree, besta2024graph, kojima2022large}, including neuro-symbolic methods~\citep{hua2022system, weir2023nellie, weir2024enhancing}. Building on this, other work considers how to optimize exemplars used for in-context learning~\citep{dong2022survey} and chain-of-thought prompting, known as ``rationale refinement''~\citep{liu2021makes, fu2022complexity}. Problem-decomposition is also shown to be effective~\citep{zhou2022least, khot2022decomposed}. 

\subsection{LLM reasoning evaluation}

LLM reasoning abilities are often tested on natural language benchmarks and commonly seen problems like arithmetic~\citep{cobbe2021training, amini2019mathqa, hendrycks2021measuring}, commonsense reasoning~\citep{bhargava2022commonsense}, and other, sometimes generative, tasks~\citep{lake2018generalization, pasupat2015compositional, lin2019commongen} and task collections~\citep{srivastava2022beyond}.
LLMs have been shown to lack sufficient reasoning capability across a range of tasks including multi-step planning and complex inference~\citep{valmeekam2022large}.
\citet{fan2023nphardeval} introduce an LLM reasoning benchmark on algorithmic problems through NP-hard complexity, and \citet{hazra2024can} show that LLMs struggle to complete simple 3SAT problems.
\citet{patel2021nlp} demonstrate that much of LLM mathematical reasoning can be explained by shallow heuristics, \citet{razeghi2022impact} similarly find that term frequency in training data impacts models' in-context learning ability, and \citep{mccoy2023embers} theorizes that ``embers'' of autoregression are polluting non-autoregressive task performance.

The effect of novelty on performance has also been explored in prior work, generally via investigating perturbations of existing tasks, taking existing problems (that are often inherently quite complex, and are only easy because they are well known, e.g., addition of numbers or logical reasoning over natural language) and changing one small aspect of the problem \citep{wu2024reasoning,saparov2023testing}.
We also study task novelty but push both language simplicity and language unfamiliarity to their limits, by exploring simple languages recognized by randomly sampled DFAs. This enables us to best isolate the power of ICL in the language domain, where LLMs should perform best.

\subsection{Training transformers on Formal Languages}

A key assumption behind this work is that the tasks we are using to evaluate LLMs are solvable by LLMs.
\citet{vafa2025evaluating} frame world modeling (a statistical model inferring the true underlying causal graph behind the data being observed) as a latent DFA identification task, finding that transformers trained on DFA traces (of massive DFAs representing board games and city maps) do not reconstruct the underlying DFA. Other work also trains language models on formal languages \citep{butoi2024training,bhattamishra2023understanding,valvoda2022benchmarking} and probabilistic formal languages \citep{borenstein2024languages}. \citet{akyurek2024context} find that transformers trained on 4-12 state DFA transducer traces more effectively learn to in-context-learn regular languages than RNNs or $n$-\textsc{Gram} models. Therefore, in this work, where we evaluate much larger LLMs on much simpler 3-state DFAs, we can be confident that underperformance relative to $n$-\textsc{Gram}s is not linked to inherent transformer limitations and must be instead related somehow to specific properties of foundation models.
\section{DFA Reasoning Tasks}

\subsection{DFAs and Regular Languages}

\begin{wrapfigure}{R}{0.6\textwidth}
\vspace{-6mm}
\includegraphics[width=1.\linewidth]{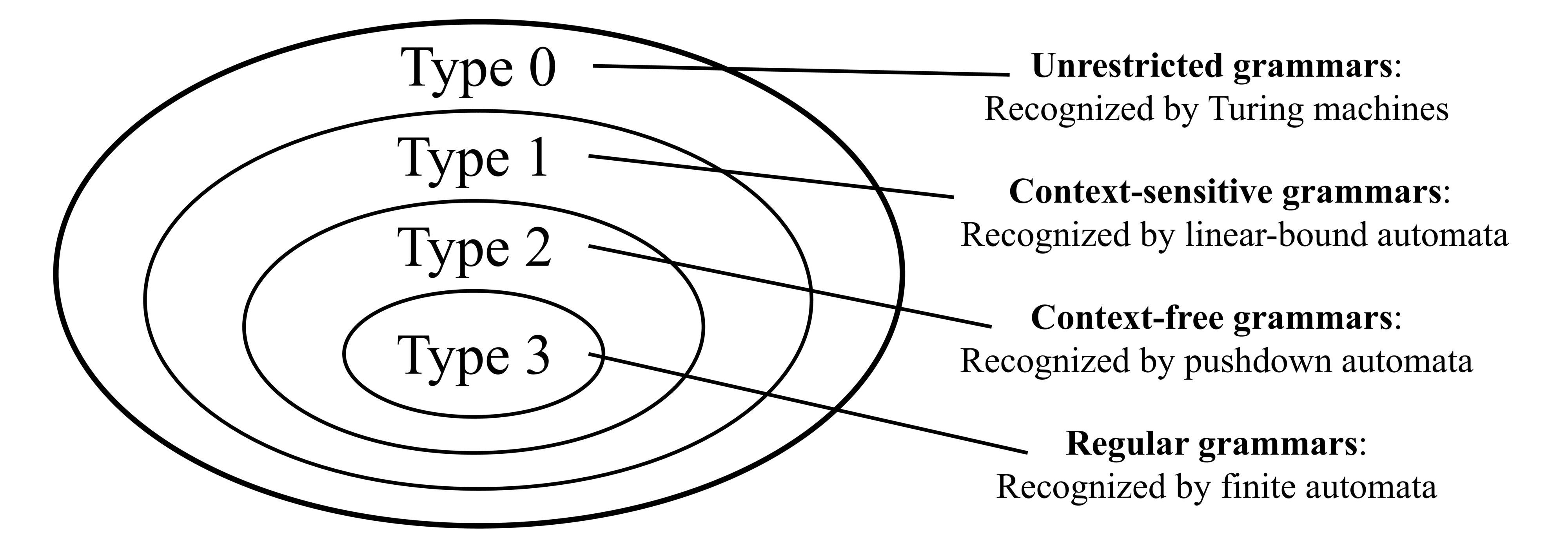}
\caption{An illustration of Chomsky's hierarchy of languages, ranging from Type 0 to Type 3, which are defined by what formal models can recognize their grammars. In this work, we focus on the simplest language type in the hierarchy, regular grammars, which are recognized by deterministic finite automata (DFAs).}
\label{fig:chomsky}
\vspace{-10mm}
\end{wrapfigure}

The original Chomsky Hierarchy~\citep{chomsky1959certain} separates language into four types (Figure~\ref{fig:chomsky}). We focus on the task of understanding Type 3 languages, the simplest form of language in the hierarchy, that are recognized by a Deterministic Finite Automata (DFAs) whose outputs are boolean (\{0, 1\}). Examples of languages recognized by DFAs include 
simple ones like \texttt{binary strings with an even number of ones}, and even such examples as \texttt{numbers in base 10 divisible by 7}. Type 3 languages are also known as regular languages, which are recognized by regular expressions.

One simple metric of the difficulty of a regular language is the number of states in the corresponding DFA, i.e., the amount of working memory.\footnote{There are other metrics of difficulty, but we choose number of states as it is highly parsimonious.} 2-state DFAs have the property that their set of states is no larger than the output set $\{0, 1\}$, and, therefore, do not have any hidden state. We thus explore 3-state DFAs, as this is the simplest nontrivial case.

\subsection{Language Reasoning Tasks}

We define a language reasoning task as a task corresponding to some latent language $\mathcal L$, where a set of positive/negative examples is provided to a model, with the goal being either classification of a new string in this language, or completion of an existing string to place it in the language.

\subsubsection{Sequence Completion Task}

We first pose a \emph{sequence completion} task, in which models must complete a sequence in a given DFA's language. This mirrors how foundation models are trained using masked language modeling, where data is presented in this format, with several \emph{example sequences} in a given language followed by a \emph{distinct prefix} that needs to be completed.

To generate test cases for this task given a DFA, we (1) sample 30 example sequences of length 10 that this DFA accepts, and then (2) sample a distinct prefix of length 5 that is not a prefix of any of our 30 example sequences, with the property that there exists some length-$\leq 5$ \emph{completion} of this prefix that the DFA would accept. The task is to find a completion (not necessarily the same completion found in sampling) of this prefix of between 1 and 5 characters such that the DFA accepts the full sequence. For details on sampling, see Appendix~\ref{app:sampling-of-sequence-completion-tasks}.

We evaluate models by (1) sampling a DFA, (2) sampling 30 problem instances at random (each of which contains 30 example sequences and a distinct prefix), and then (3) computing a binary prediction score (whether or not the predicted completion creates a valid string in the language) for each instance separately, then computing a correctness metric as a fraction. We then average this metric over several sampled DFAs to produce our accuracy score.

\subsubsection{Transducer Task}

\label{sec:sequence-completion-bad}
While the sequence completion task is the natural one that comes to mind as a basic language task, it has a difficulty-gap problem. Specifically, many DFAs, including the one shown in Figure~\ref{fig:intro-example}, recognize languages that are particularly difficult to identify based on a set of examples, unless you build some kind of world model.\footnote{
The difficulty gap exists because a set of recognized sequences of length 10 gives no direct insight into intermediate states between the first and tenth token. As such, to be able to utilize this information for languages like the one in Figure~\ref{fig:intro-example} where there are no ``resets'' (sequences of symbols that necessarily lead to a particular state), a model must be capable of hollistically evaluating the entire sequence, probably requiring a world model. Many other DFAs contain these resets, but do so in such a way that makes it possible to e.g., recognize that all sequences that end in \texttt{a} are in the language, making the problem trivial.
}
This is problematic as we would like to be able to assess the performance of language models at pattern recognition, independent of their world modeling abilities. To assess pattern recognition, we explore the Transducer task.

In this task, an input sequence is annotated with an output at each token, the final output is masked, and the masked output is predicted by a language model.
E.g., given the language \texttt{even number of 'a' tokens} and the input \texttt{abcabcaabbccaa}, the annotated string (all that is provided to the model) is \texttt{a0b0c0a1b1c1a0a1b1b1c1c1a0a} and the output to predict is \texttt{1}. For each problem instance, we provide 30 symbols, and for the first 29, the corresponding transducer output.

This task is significantly more transparent than the sequence completion task as the model has access to intermediate outputs, an (imperfect) proxy for intermediate state.

\subsection{Baselines}

To contextualize LLM accuracies, we provide several baselines with varying degrees of sophistication.

\paragraph{Sequence Completion Task} For the Sequence Completion task, we have four baselines.
\begin{itemize}
    \item \textsc{Random}$_S$ baseline: produce a random string of length 5 characters. While this might seem redundant as it should have a success rate of 50\%, in practice our rejection sampling approach (see Appendix~\ref{app:sampling-of-sequence-completion-tasks}) leads to a slight bias towards DFAs with more accept states. This baseline measures that bias.
    \item \textsc{Common-Suffix}$_S$ baseline: find the completion $s$ of length between 1 and 5 that maximizes $($\# of occurrences as a suffix${} \times |s|)$. This baseline does not take the distinct prefix into account, and instead tries to find a universal completion that will always end in an accept state for this language.
    \item $n$-\textsc{Gram}$_S$ baseline: we take the last $n - 1$ characters of the distinct prefix and search to see if they appear in any of the example sequences at a position where the sequence following is an appropriate length to be a completion (at least 1 but at most 5). We then take a plurality vote among the completions and return this, breaking ties arbitrarily. If there are no matches, we return the result of $(n - 1)$-\textsc{Gram}$_S$. Technically these cover more than $n$ characters, since the completion is often $>1$ character long; for simplicity, however, we keep the naming consistent with the Transducer baselines. Despite the similarity between an $n$-\textsc{Gram} and a DFA in terms of token-to-token transitions, $n$-\textsc{Grams} do not have access to DFA hidden state and thus cannot solve arbitrary DFA language problems, regardless of $n$.
    \item \textsc{Brute-Force}$_S$: take all possible DFAs with 3 states and 3 symbols. Filter for ones that accept all the example sequences. Then try all remaining DFAs on all $3^5$ possible 5-length completions and return the completion that the maximal number of DFAs accept, breaking ties arbitrarily.
\end{itemize}

Note that these baselines are entirely unparameterized and operate identically regardless of the underlying DFA. This makes them direct comparisons to using LLMs in in-context-learning\footnote{One thing to note is that LLMs are required to determine that they are performing next token prediction on a particular string from a natural language description such as ``You are a sequence completion model\ldots,'' while $n$-\textsc{Gram} models are programmed to do so. However, we believe all LLMs we evaluate are sophisticated enough to accomplish this without issue.}. We do not consider \textsc{BruteForce}$_S$ to be a reasonable comparison due to its computational complexity, and instead consider it an upper bound on performance on this particular task. We choose $n$-\textsc{Gram} baselines as they are are unambiguously representable by transformers~\citep{svete2024transformers}, so a transformer model should be able to match their performance.

\paragraph{Transducer Task}  We have similar baselines for the Transducer task.
\begin{itemize}
    \item \textsc{Null}$_T$ baseline: for a given DFA, whichever of the following strategies produces a higher accuracy: always predict 0 or always predict 1.
    \item $n$-\textsc{Gram}$_T$ baseline: take the $n - 1$ symbols ending at the end of the concatenated transducer sequence (e.g., for $n=5$ and the above example, this would be \texttt{1a0a}). If that sequence does not appear elsewhere in the sequence, return the result of the $(n-1)$-\textsc{Gram}$_T$ baseline. Otherwise, take the token that appears immediately after each occurrence. If there is a majority, return that, otherwise return the last example.
    \item \textsc{BruteForce}$_T$: take all possible DFAs with 3 states and 3 symbols. Filter them for ones that match the given transducer sequence. Take this set and predict the next token. Take a majority vote among these, returning 1 by default if there is no majority.
\end{itemize}

\section{Experiments}

We evaluated the open-weight models Llama 3-8B, Llama 3-70B \citep{llama38B}, Llama 3.1-8B \citep{llama3.18B}, Llama 3.1-8B-Instruct \citep{llama3.18BInstruct}, Llama 3.1-70B \citep{llama3.170B}, Mistral Nemo Minitron 8B \citep{mistralnemominitron8B}, Mistral Nemo Base 2407 \citep{mistralnemobase2407} and Mistral Nemo Instruct 2407 \citep{mistralnemoinstruct2407}, Gemma 7B \citep{gemma7b}, Falcon 7B \citep{falcon40b}, Qwen 2.5-7B and Qwen 2.5-32B \cite{team2024qwen2}.

We also evaluated the open-weight code models StarCoder2-15B \citep{lozhkov2024starcoder}, Codestral-22B-v0.1 \citep{Codestral22Bv0.1}, Deepseek Coder 33B Instruct \citep{DeepseekCoder33bInstruct}, Qwen2.5-Coder-7B, Qwen2.5-Coder-7B-Instruct, and Qwen2.5-Coder-32B-Instruct \citep{hui2024qwen2}.

Finally, we evaluated the proprietary models Claude 3.5 Sonnet \citep{claude3.5sonnet}, GPT-3.5-turbo-instruct, GPT-3.5 Chat (turbo-0125) \citep{gpt3.5turbo}, GPT-4o-mini (2024-07-18), GPT 4o (2024-05-13) \citep{openai2024gpt4ocard}, o3-mini (2025-01-31) \citep{o3minicard}, and gpt-5 (2025-08-07) \citep{gpt5card}.

For both tasks, we consider two main prompting formats. \textsc{Basic} provides no context, presenting the problem as a generic sequence generation or next-token prediction task, where output is provided immediately following the input, with no space to think. \textsc{Basic-COT} provides the same prompt but asks the model to think step by step and provide an answer. These prompts test ICL, presenting the task in an unstructured manner and requiring the model to learn the problem structure via induction. Our main results are the maximum over these two prompting strategies.

We also provide three ``control'' prompting formats where information on the problem structure is provided. \textsc{More-Expl} explains that the strings are generated from a simple grammar, but is otherwise identical to \textsc{Basic}. This remains a sequence generation/next token prediction task. \textsc{DFA-COT} provides the full structure of the latent language, stating that it is a 3-state DFA, and additionally invokes chain-of-thought reasoning to help the model reason over the task. \textsc{Red-Green} casts the tasks as independent word problems that describe the underlying grammar structure without relying on world knowledge about DFAs and regular languages. It describes an N-state DFA as a house with N rooms, each of which has 3 portals that deterministically go to other rooms (or back to the same room), where the walls of each room are red or green (mirroring transducer output symbols 0 and 1). Similarly to \textsc{DFA-COT}, the model is given space to show work before providing a tagged answer.

We produce versions of each of these prompts for each task, denoting these with a subscript ${}_S$ for sequence completion prompts and ${}_T$ for transducer prompts. Full listings of these prompts can be found in Appendix~\ref{app:prompt-listings}. While no finite set of prompts will be fully sufficient to capture all possible model behavior, we believe maximizing over both \textsc{Basic} prompts allows both models that perform best at next-token-prediction and those that perform better in chain-of-thought reasoning to do their best.

For each open weight model, we used a local VLLM \citep{kwon2023efficient} server for evaluation and always evaluated on 1000 distinct DFAs. For GPT-4o and Claude, o3-mini, and gpt-5, we evaluated on 30 DFAs due to computation costs. (Due to greater interest in o3-mini's performance on \textsc{Red-Green}$_T$, we used 100 to get a more precise estimate). For gpt-3.5 and gpt-4o-mini, we evaluated on 100 DFAs. All models were evaluated with temperature 0, except reasoning models o3-mini and gpt-5, which do not support a custom temperature.

\newcommand{\tablesize}{8pt}

\begin{table*}[htbp]
    \centering
    \fontsize{\tablesize}{\tablesize}\selectfont
    {\renewcommand{\arraystretch}{1.25}\begin{tabular}{l|ccc|ccccc}
\hline
\bf Model & \bf Size & \bf IT? & \bf Code? & \bf Sequence Completion & \bf SR & \bf Transducer & \bf TR\\
\hline
\multicolumn{8}{c}{ \bf Baselines} \\
\hline
\textsc{BruteForce} & -- &  &  & \cellcolor{lightgray} 100.0 (99.9--100.0) & 1 & \cellcolor{lightgray} 96.4 (96.2--96.7) & 1 \\
\hline
6-\textsc{Gram} & -- &  &  & \bf 91.7 (91.0--92.4) & 2 & \bf 93.5 (93.1--93.9) & 2 \\
\hline
5-\textsc{Gram} & -- &  &  & 91.2 (90.4--91.9) & 3 & 93.4 (93.0--93.7) & 3 \\
\hline
4-\textsc{Gram} & -- &  &  & 89.6 (88.7--90.4) & 4 & 91.1 (90.6--91.6) & 4 \\
\hline
3-\textsc{Gram} & -- &  &  & 87.0 (86.1--87.8) & 5 & 87.0 (86.4--87.6) & 19 \\
\hline
2-\textsc{Gram} & -- &  &  & 83.3 (82.2--84.2) & 7 & 74.5 (73.6--75.3) & 30 \\
\hline
\textsc{Common-Suffix} & -- &  &  & 84.7 (83.6--85.6) & 6 & -- & -- \\
\hline
\textsc{Random}$_S$/\textsc{Null}$_T$ & -- &  &  & 53.3 (51.7--54.7) & 32 & 68.9 (68.2--69.6) & 31 \\
\hline
\multicolumn{8}{c}{ \bf Open Weight Completion} \\
\hline
llama3-8B & 8.0B &  &  & 73.8 (72.4--75.1) & 22 & 87.5 (86.9--88.0) & 18 \\
\hline
llama3-70B & 70.6B &  &  & 71.4 (70.0--72.7) & 29 & 87.7 (87.2--88.3) & 15 \\
\hline
llama3.1-8B-Instruct & 8.0B & \checkmark &  & 75.3 (74.0--76.6) & 19 & 85.9 (85.3--86.5) & 22 \\
\hline
llama3.1-8B & 8.0B & \checkmark &  & 75.2 (73.8--76.3) & 20 & 88.0 (87.5--88.6) & 10 \\
\hline
llama3.1-70B & 70.0B & \checkmark &  & 71.8 (70.4--73.1) & 28 & 87.7 (87.2--88.2) & 17 \\
\hline
qwen-2.5-7B & 7.6B &  &  & 73.5 (72.1--74.8) & 24 & \bf 88.7 (88.2--89.2) & 5 \\
\hline
qwen-2.5-32B & 32.5B &  &  & 76.8 (75.5--78.0) & 16 & 88.3 (87.8--88.8) & 7 \\
\hline
mistral-nemo-minitron-8B & 8.4B &  &  & \bf 78.7 (77.5--79.8) & 13 & 88.6 (88.0--89.1) & 6 \\
\hline
mistral-nemo-base-12B & 12.2B &  &  & 75.5 (74.3--76.6) & 18 & 87.9 (87.4--88.4) & 13 \\
\hline
mistral-nemo-instruct-12B & 12.2B & \checkmark &  & 72.2 (70.9--73.4) & 27 & 88.0 (87.5--88.5) & 11 \\
\hline
gemma-7b & 8.5B &  &  & 72.6 (71.3--73.7) & 25 & 82.1 (81.4--82.7) & 27 \\
\hline
falcon-7b & 7.2B &  &  & 69.0 (67.6--70.2) & 30 & 84.9 (84.3--85.5) & 24 \\
\hline
\multicolumn{8}{c}{ \bf Open Weight Code} \\
\hline
starcoder2-15b & 16.0B &  & \checkmark & 73.5 (72.0--74.7) & 23 & 87.7 (85.8--89.5) & 16 \\
\hline
codestral-22B & 22.2B &  & \checkmark & 78.0 (76.8--79.1) & 14 & 86.6 (86.0--87.1) & 21 \\
\hline
deepseek-coder-33b-instruct & 33.3B & \checkmark & \checkmark & 76.7 (75.3--77.8) & 17 & 85.6 (85.0--86.2) & 23 \\
\hline
qwen-2.5-coder-7B & 7.6B &  & \checkmark & \bf 79.5 (78.4--80.5) & 10 & 88.2 (87.6--88.7) & 9 \\
\hline
qwen-2.5-coder-instruct-7B & 7.6B & \checkmark & \checkmark & 79.5 (78.3--80.5) & 11 & \bf 88.3 (87.8--88.8) & 8 \\
\hline
qwen-2.5-coder-instruct-32B & 32.8B & \checkmark & \checkmark & 79.2 (78.0--80.3) & 12 & 87.9 (87.4--88.4) & 12 \\
\hline
\multicolumn{8}{c}{ \bf Proprietary} \\
\hline
gpt-3.5-instruct & ? & \checkmark &  & 67.3 (63.1--71.5) & 31 & \bf 87.8 (85.9--89.6) & 14 \\
\hline
gpt-3.5-chat & ? & \checkmark &  & N/A & -- & 66.8 (63.4--69.8) & 32 \\
\hline
gpt-4o-mini & ? & \checkmark &  & 72.4 (68.1--76.3) & 26 & 79.8 (77.3--82.2) & 28 \\
\hline
gpt-4o & ? & \checkmark &  & 74.8 (69.3--80.4) & 21 & 83.7 (80.1--86.9) & 25 \\
\hline
claude-3.5 & ? & \checkmark &  & \bf 82.8 (77.5--87.5) & 8 & 86.9 (83.3--90.0) & 20 \\
\hline
o3-mini & ? & \checkmark &  & 81.1 (76.0--85.8) & 9 & 74.7 (70.7--78.8) & 29 \\
\hline
gpt-5 & ? & \checkmark &  & 77.9 (71.6--84.0) & 15 & 83.6 (79.9--87.1) & 26 \\
\hline
\end{tabular}
}

    \caption{Results for our experiments on both the Transducer and Sequence completion tasks. Each cell contains the mean performance across DFAs for the best-performing \textsc{Basic} prompt (see Table~\ref{tab:results-multiprompt} for details), with 95\% confidence intervals of the mean in parentheses. ``N/A'' is used whenever the model returned an invalid result at least 25\% of the time. (IT = Instruction-Tuned, TR/SR = rank of the given model on Transducer/Sequence Completion.)}
    \label{tab:results-main}
\end{table*}

\begin{table*}[htbp]
    \fontsize{\tablesize}{\tablesize}\selectfont
    \centering
    {\renewcommand{\arraystretch}{1.25}\begin{tabular}{l|cc|ccc}
\hline
\bf Model & \bf \textsc{Basic} & \bf \textsc{Basic-COT} & \bf \textsc{More-Expl} & \bf \textsc{DFA-COT} & \bf \textsc{Red-Green} \\
\hline
\multicolumn{6}{l}{ \bf Sequence Completion} \\
\hline
gpt-4o-mini & \bf 72.4 (68.1--76.3) & 60.0 (55.8--64.4) & 70.5 (66.4--74.6) & 58.0 (53.4--62.4) & 59.1 (54.9--63.2) \\
\hline
gpt-4o & 72.1 (65.9--78.2) & \bf 74.8 (69.3--80.4) & N/A & 67.4 (60.8--73.8) & 74.4 (69.9--78.6) \\
\hline
claude-3.5 & N/A & 82.8 (77.5--87.5) & N/A & \bf 84.0 (79.3--88.4) & 80.0 (74.9--85.2) \\
\hline
o3-mini & N/A & \bf 81.1 (76.0--85.8) & N/A & 58.2 (49.6--66.8) & 69.8 (64.4--75.0) \\
\hline
gpt-5 & 77.9 (71.6--84.0) & 75.1 (68.6--81.7) & 68.9 (61.0--76.7) & 66.0 (58.9--72.8) & \bf 87.5 (83.1--91.5) \\
\hline
\multicolumn{6}{l}{ \bf Transducer} \\
\hline
gpt-4o-mini & \bf 79.8 (77.3--82.2) & 66.5 (64.3--68.7) & 76.7 (74.2--79.3) & 65.2 (63.1--67.4) & 74.5 (72.0--77.0) \\
\hline
gpt-4o & \bf 83.7 (80.1--86.9) & 67.8 (62.4--73.2) & 82.6 (79.1--85.9) & 67.8 (63.1--72.3) & 82.6 (78.8--86.3) \\
\hline
claude-3.5 & 86.9 (83.3--90.0) & 74.2 (70.0--78.3) & \bf 87.1 (83.9--90.2) & 76.4 (72.9--79.9) & 82.9 (78.9--86.9) \\
\hline
o3-mini & 72.8 (68.4--77.3) & 74.7 (70.7--78.8) & 74.7 (70.3--79.2) & 86.1 (83.9--88.4) & \bf 92.4 (91.3--93.5) \\
\hline
gpt-5 & 83.6 (79.1--87.7) & 83.6 (79.9--87.1) & 85.2 (81.0--88.8) & \bf 96.7 (95.3--98.0) & 96.6 (95.4--97.8) \\
\hline
\end{tabular}
}

    \caption{Results for models where we investigated multiple prompts (we only used \textsc{Basic} on other models). We bold the best prompt for each model. Non-COT prompts consistently work better for the Transducer task, with more mixed results on sequence completion.}
    \label{tab:results-multiprompt}
\end{table*}

\section{Results}

Main results for all tasks are presented in Table~\ref{tab:results-main}. We ignore non-answers, i.e., if for a given DFA a model gets 25 correct answers, 1 incorrect answer, and responds with an unparseable result on 4, this counts as a 25/26, not a 25/29. We then aggregate across DFAs. All comparisons involving 4-\textsc{Gram}, 5-\textsc{Gram}, and 6-\textsc{Gram} to all other models are statistically significant (see Appendix~\ref{app:significance} for details).

\subsection{Sequence Completion}

As seen in Table~\ref{tab:results-main}, this task is nearly always fully determined, that is, it can be solved with $\sim$100\% accuracy in theory, as demonstrated by \textsc{BruteForce}$_S$ results. Of course, \textsc{BruteForce}$_S$ is extremely computationally expensive, and, as such, we primarily focus on the $n$-\textsc{Gram}$_S$ heuristics as our baselines. Still, we find that $n$-\textsc{Gram}$_S$ heuristics tend to outperform LLMs.

As seen in Table~\ref{tab:results-multiprompt}, we find that giving the model the opportunity to logically reason about the prompt via chain-of-thought and present a conclusion has inconsistent results. Specifically, we find that for \texttt{gpt-4o-mini}, immediately predicting a next token seems to be better, while for \texttt{gpt-4o} and \texttt{gpt-5} there is no large effect. \texttt{claude-3.5} and \texttt{o3-mini} are unable to answer the \textsc{Basic}$_S$ prompt at all, but outperform other proprietary models when using \textsc{Basic-COT}$_S$. In this task, revealing the problem structure appears to not have a massive effect on performance, with \texttt{gpt-5} being the only model to incorporate this information into a statistically significantly improved performance, and even then only in the \texttt{Red-Green}$_S$ word problem prompt (still underperforming 4-\textsc{Gram}$_S$).

Additionally, we find that in this task, code-specific open-weight models tend to perform better than sequence completion models, suggesting some generalized ability to produce strings from novel languages demonstrated by example. Overall, the relative performances of LLMs and prompts generally comport to heuristics on which models and prompting strategies should work best (with the notable exception of gpt-5). Nonetheless, LLMs underperform simple $n$-\textsc{Gram} heuristics.

One potential problem with using this task for cross-model comparisons is the relevance of tokenization. We found that forcing uniform tokenization by using commas in the prompt uniformly reduced accuracy, see Appendix~\ref{app:sequence-completion-commas} for details; we confirmed that the LLMs we investigate could reason about strings without commas, see Appendix~\ref{app:results-tokenization-test-regexp} for details.

\subsection{Transducer}

Unlike sequence completion, this task is not fully determined, with $\textsc{BruteForce}_T$ getting 96.4\% accuracy. Comparisons are still valid as all models see the same fraction of unsolvable instances.

We find that in general all LLMs underperform a 4-\textsc{Gram}$_T$ model, demonstrating that they are unable to adequately solve this task. The relative performance of the models also does not correspond to their overall scale, with open weight LLama-3 and Mistral Nemo 8B parameter models outperforming much larger proprietary models. Even within a model class we find no clear pattern: all other GPT models are outperformed by GPT 3.5, Llama 3-70B has similar performance to Llama 3-8B, and the Mistral Nemo 12B models perform similarly to Nemo Minitron 8B. Coding models also demonstrate no advantage on this task.

The generally lower performance of chat-oriented models suggests this task is better suited to non-chat models. More specifically, as seen in Table~\ref{tab:results-multiprompt}, our \textsc{Basic-COT}$_T$ prompt results in underperformance by all non-reasoning models, suggesting that models are generally most able to solve this task when it is a simple next-token-prediction task. Providing the problem structure also does not help non-reasoning models improve substantially, but does allow \texttt{o3-mini} to perform well, and \texttt{gpt-5} to completely solve the task (achieving parity with \textsc{BruteForce}), demonstrating that reasoning models' underperformance at the original ICL language reasoning task is not due to the underlying difficulty of the task itself.

We conclude that LLM ICL is unable to perform well at language inference. This failure cannot be attributed to a lack of world modeling, as $n$-\textsc{Gram}$_T$ models do not construct world models. Instead, it seems the LLMs are unable to detect patterns when those patterns are drawn from an unfamiliar and unknown source, even a relatively simple one.

\subsection{Comparison of Benchmarks}

\begin{figure*}[htb]
    \centering
    \includegraphics[width=\textwidth]{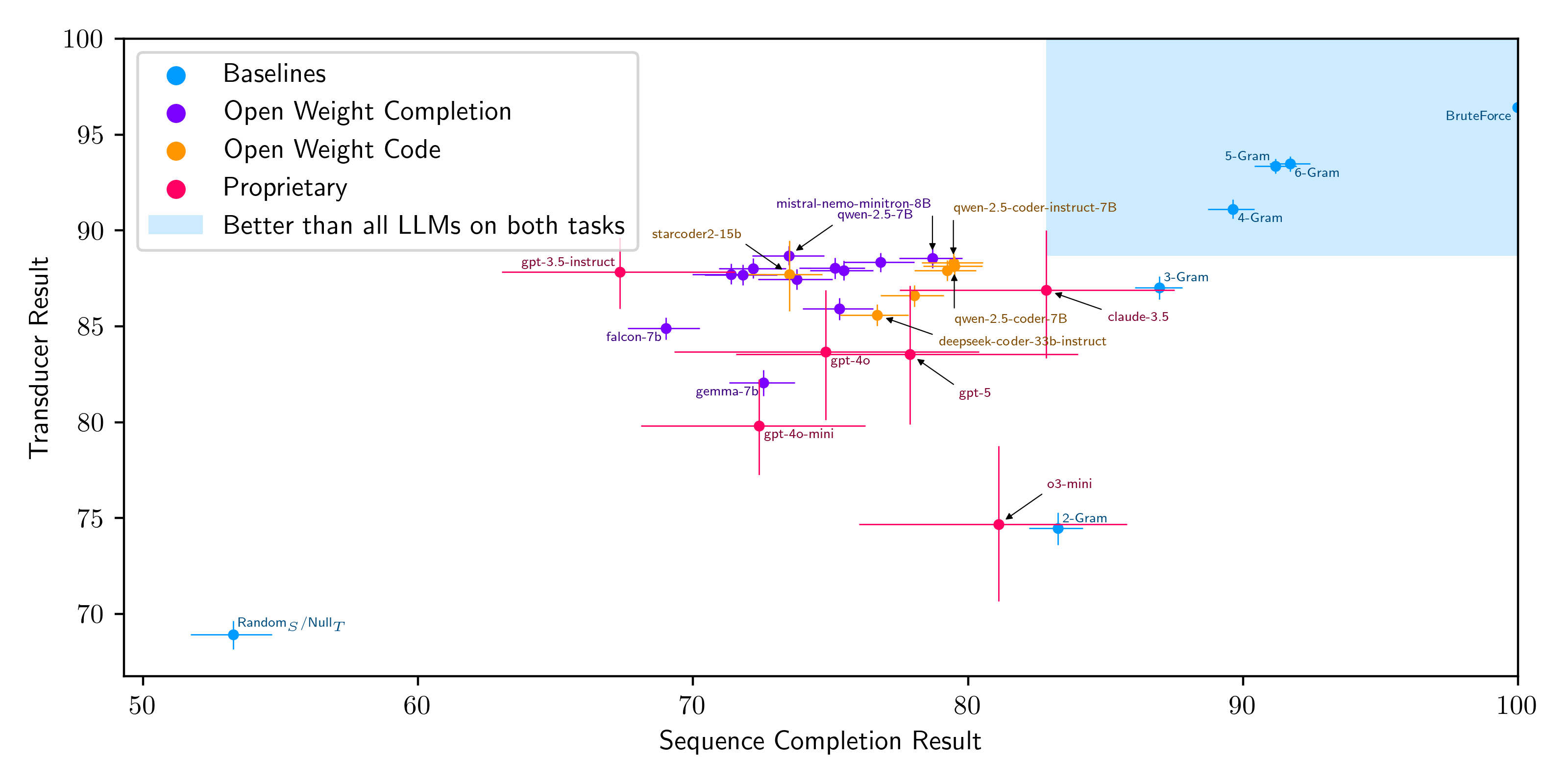}
    \caption{
        Transducer and sequence completion results plotted against each other. Points are the mean over several DFAs, with 95\% confidence intervals. Points are colored by model type, with the best and worst model by each metric in each category labeled, as well as all baseline \& proprietary models.
    }
    \label{fig:sc-vs-transducer}
\end{figure*}

Figure~\ref{fig:sc-vs-transducer} displays the relationship between model performance on the Sequence Completion and Transducer benchmarks. While at a high level, there is a positive correlation between the two, there are a few notable differences. For one, the Code models perform notably better than other open weight models on Sequence Completion, but not on Transducer. Additionally, on Transducer, a ceiling on performance is observed, where non-reasoning LLMs cluster together between 3-\textsc{Gram}$_T$ and 4-\textsc{Gram}$_T$ performance; this clustering does not appear on the Sequence Completion benchmark.

\section{Conclusion}

Our findings highlight significant weaknesses in large language models' ability to in-context-learn entirely novel language reasoning problems, even simple ones solely involving next-token prediction on basic languages recognized by 3-state DFAs. These results, combined with that of previous work demonstrating that large language models can quite accurately perform a variety of language tasks, suggests that LLMs solve language problems via a mechanism distinct from general language reasoning ability. Our use of n-gram baselines and next-token prediction tasks allows us to exclude the possibility that the issue is primarily related to LLMs'  lack of world modeling or any inherent limitations of next-token prediction models. We believe our results suggest that LLMs have learned individual models of particular languages, but not a general theory of language.


Interestingly, in our transducer experiments, directly predicting the next token rather than explicitly reasoning through the problem works better, except for reasoning models where they perform similarly.
While our conclusions are limited by the finite nature of our prompt set, this suggests that they do, in fact, possess some latent understanding of language, but this understanding is inferior to basic n-gram models for $n > 3$.


One potential goal for foundation models is to replace all machine learning with ICL. Our results suggest that current models are not progressing towards this goal.
\section*{Impact Statement}

Aside from the social consequences of this work as related to advancing the field of Machine Learning in general, this work has the goal of advancing the field of benchmarks in Machine Learning. While we view this as a positive objective, as it ensures that models are being evaluated fairly, it might have negative consequences insofar as benchmarking techniques might be best left unpublished to prevent deliberate or unintentional overfitting.

\ifdefined\isaccepted
\section*{Acknowledgments}
\small
We thank Marc Marone and Theo Olausson for their advice on experiment design.
\fi



\bibliography{main}
\bibliographystyle{neurips}

\newpage
\appendix
\onecolumn
\section{Details on Sampling}

\subsection{Sampling of DFAs} \label{app:sampling-of-dfas}

We use rejection sampling to sample DFAs. Specifically, we uniformly sample a start state, then for each (source state, symbol) pair, we sample a post-transition state. We also randomly assign each state to be accept or reject with probability 50\%. We then reject any DFA that has all accept or all reject states (so only DFAs with 1 or 2 accept states are allowed), or for which certain states are unreachable from the start state.

\subsection{Sampling of Sequence Completion Tasks} \label{app:sampling-of-sequence-completion-tasks}

To sample a sequence completion task, we first sample a DFA as described in Appendix~\ref{app:sampling-of-dfas}.

To sample a task instance, we sample example sequences and distinct prefix. Each example sequence is sampled uniformly from the space of $\{a, b, c\}^{10}$ and then rejected if the DFA does not accept the sequence. Our distinct prefix and completion are sampled uniformly from $\{a, b, c\}^{5} \times \{a, b, c\}^{5}$, and are rejected if the DFA does not accept the concatenation of the two, or if the prefix is the prefix of any of the previous sequences. We then discard the completion. If we, at any point, reject 50 sequences when attempting to sample a sequence or prefix, we return an error.

We run a ``pilot'' sampling for a DFA to ensure that it is valid, in which we sample an instance as described above. If there is an error in sampling this pilot instance, we reject the DFA. Otherwise, we proceed to sample our task instances. At this stage, if there is an error in sampling, we reject the instance rather than the DFA. This pilot sample rejection procedure leads to a slight bias towards 2-accept state DFAs over 1-accept state DFAs, as measured by the $\textsc{Random}_S$ baseline.

\subsection{Sampling of Transducer Tasks} \label{app:sampling-of-transducer-tasks}

We sample a DFA as described in Appendix~\ref{app:sampling-of-dfas}, and then sample random sequences  (30 in our experiments) and generate transducer traces. If every transducer trace ends with a 0 or every trace ends with a 1, we reject the DFA and resample.

\clearpage
\section{Transducer results by difficulty class} \label{app:transducer-difficulty}

\begin{figure}[!htbp]
    \centering
    \includegraphics[width=\linewidth]{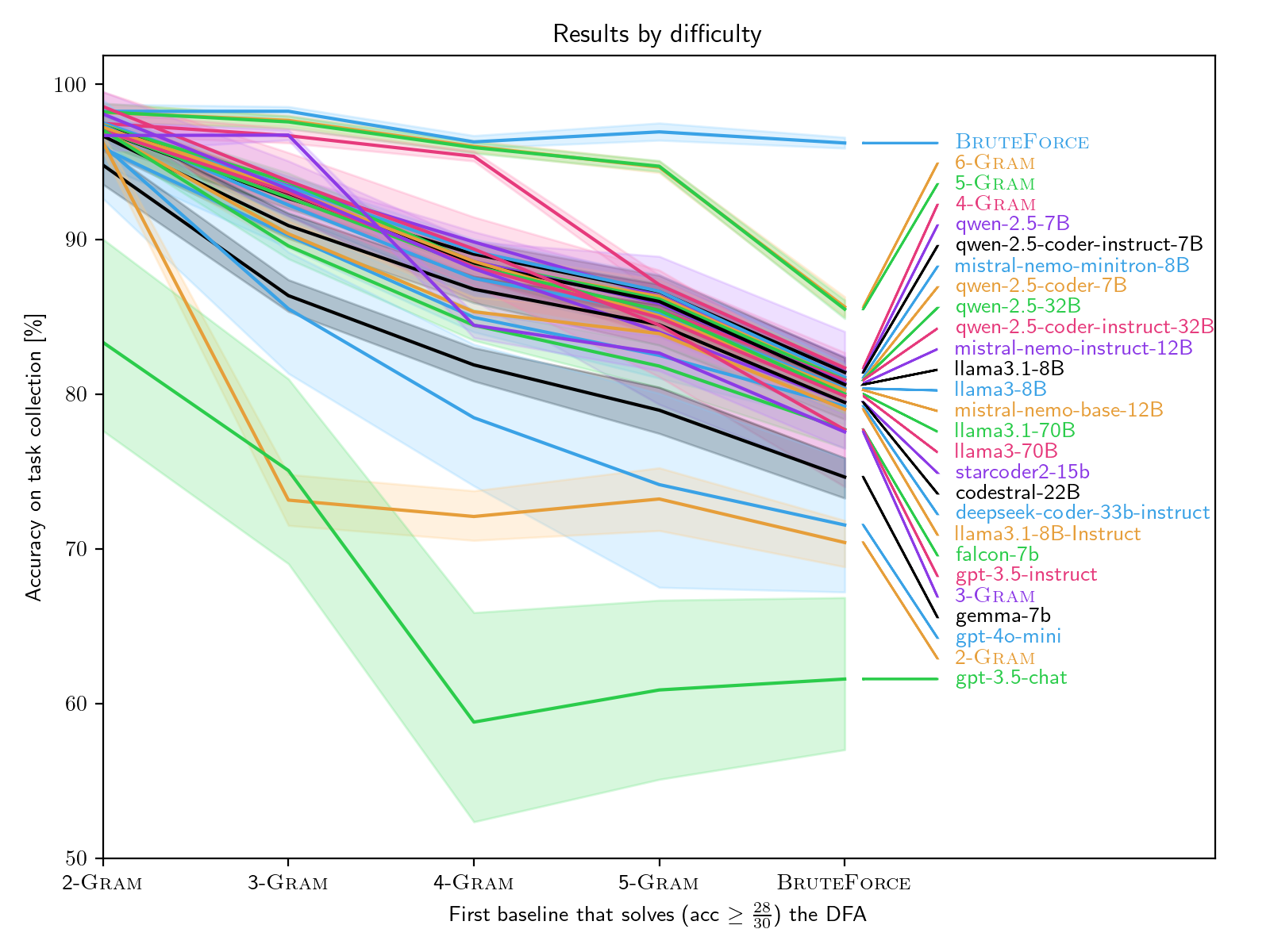}
    \caption{Transducer results by difficulty class. We classify each DFA based on which of the baselines first achieves a score of 28/30 on the given instances. 6-\textsc{Gram} is excluded as it has very similar performance to 5-\textsc{Gram}. Each model's best prompt results are plotted, with 95\% confidence intervals, for all models with at least 100 DFAs; those with 10 or 30 had error bars too large to make this analysis useful.}
    \label{fig:by-diff}
\end{figure}

Figure~\ref{fig:by-diff} displays results by difficulty level, as judged by the smallest $n$-\textsc{Gram} model that can solve a particular task. All models behave roughly monotonically, performing more poorly as difficulty increases. Additionally, we find that the best models continue to perform similarly to 4-\textsc{Gram} for tasks that 4-\textsc{Gram} does not perfectly solve.

\clearpage
\section{Case Study: Sum Modulo 3 DFA}
\label{app:case-study}

\begin{wrapfigure}{R}{0.5\textwidth}

\begin{minipage}[t]{0.1\linewidth}
{\Large (a)}
\end{minipage}
\begin{minipage}{0.8\linewidth}
\includegraphics[width=\linewidth]{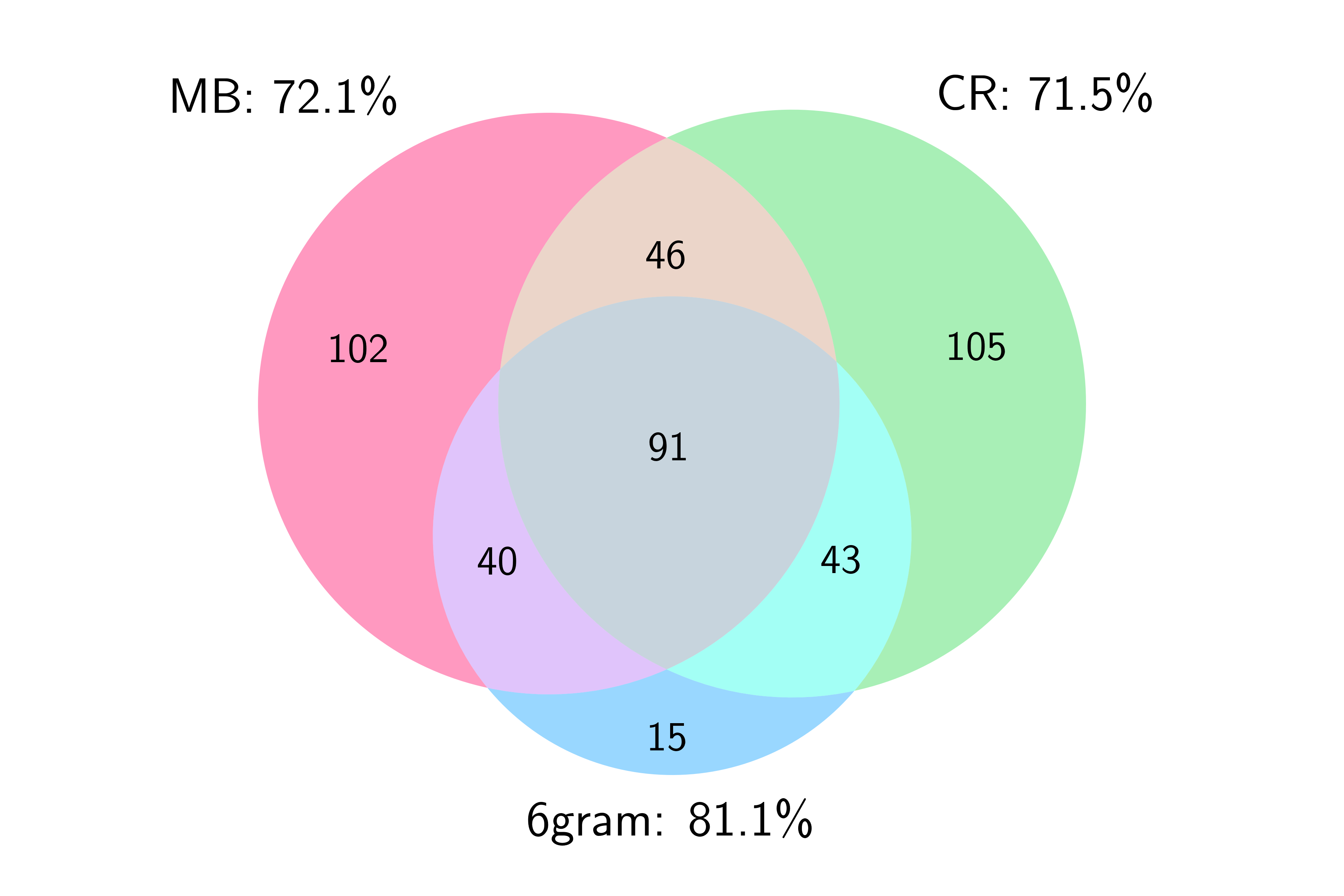}
\end{minipage}\\

\begin{minipage}[t]{0.1\linewidth}
{\Large (b)}
\end{minipage}
\begin{minipage}{0.8\linewidth}
\begin{tabular}{|r|r|r|}
\hline
 & Correct & Incorrect \\\hline
Total & 100\% & 100\% \\
\hline
\texttt{a} is no-op & 70\% & 73\% \\
\hline
\texttt{1b} and \texttt{1c} lead to 0 & 47\% & 57\% \\
\hline
2-periodic & 30\% & 47\% \\
\hline
3-periodic & 13\% & 13\% \\
\hline
2 red rooms & 7\% & 10\% \\
\hline
\end{tabular}
\end{minipage}

\caption{
Results on Sum Modulo 3 DFA.
(a) MB=mistral-nemo-minitron-8B/\textsc{Basic}$_T$, CR=claude-3.5/\textsc{Red-Green}$_T$. Venn diagram of errors (out of 1000). Labeled percentages are accuracies.
(b)~Results of qualitative analysis, out of 30 in both cases.
}
\label{fig:case-study-main}
\vspace{-5mm}
\end{wrapfigure}

We investigate the transducer task on the DFA depicted in Figure~\ref{fig:intro-example}. This DFA can be interpreted as an arithmetic check, where \texttt{a} represents 0, \texttt{b} represents 1, and \texttt{c} represents 2, and the DFA accepts strings whose sum is equal to 0 modulo 3. For this case study, we focus on the model/prompt combinations MB (mistral-nemo-minitron-8B/\textsc{Basic}$_T$: the best performing non-reasoning model combination overall) and CR (claude-3.5/\textsc{Red-Green}$_T$: the best performing non-reasoning model combination that provides an explanation, needed later for our qualitative analysis).

Figure~\ref{fig:case-study-main}a depicts the number of errors each model receives on 1000 instances of the transducer task for this DFA. Nearly all errors made by the 6-\textsc{Gram}$_T$ model were also made by at least one LLM, while the two LLMs often made unique errors. While this task is better-known than most DFAs, all 3 models perform worse on this DFA than average.

We also performed a qualitative analysis, investigating CR's outputs on the \textsc{Red-Green}$_T$ prompt to see what kind of reasoning it is using; specifically we sampled 30 examples where it had the correct answer, and 30 examples where it had the incorrect answer but the 6-\textsc{Gram}$_T$ model had the correct answer. Results of this analysis can be found in Figure~\ref{fig:case-study-main}b.
We find that, in general, CR is following a 3-\textsc{Gram} approach, learning rules relating to the conditions under which the previous output and symbol can be used to predict the next output. Specifically, it is able to learn that \texttt{a} does not change the output, and that \texttt{b} and \texttt{c} will lead a \texttt{1} state to a \texttt{0} state. These results comport with the overall finding of Table~\ref{tab:results-main}, where we found that 3-\textsc{Gram}$_T$ was the largest $n$-\textsc{Gram}$_T$ that any non-reasoning LLM outperformed, as well as our finding that LLM performance decreases for tasks that are not solvable by $n$-\textsc{Gram}s; see Appendix~\ref{app:transducer-difficulty} for details.

\begin{figure}[!ht]
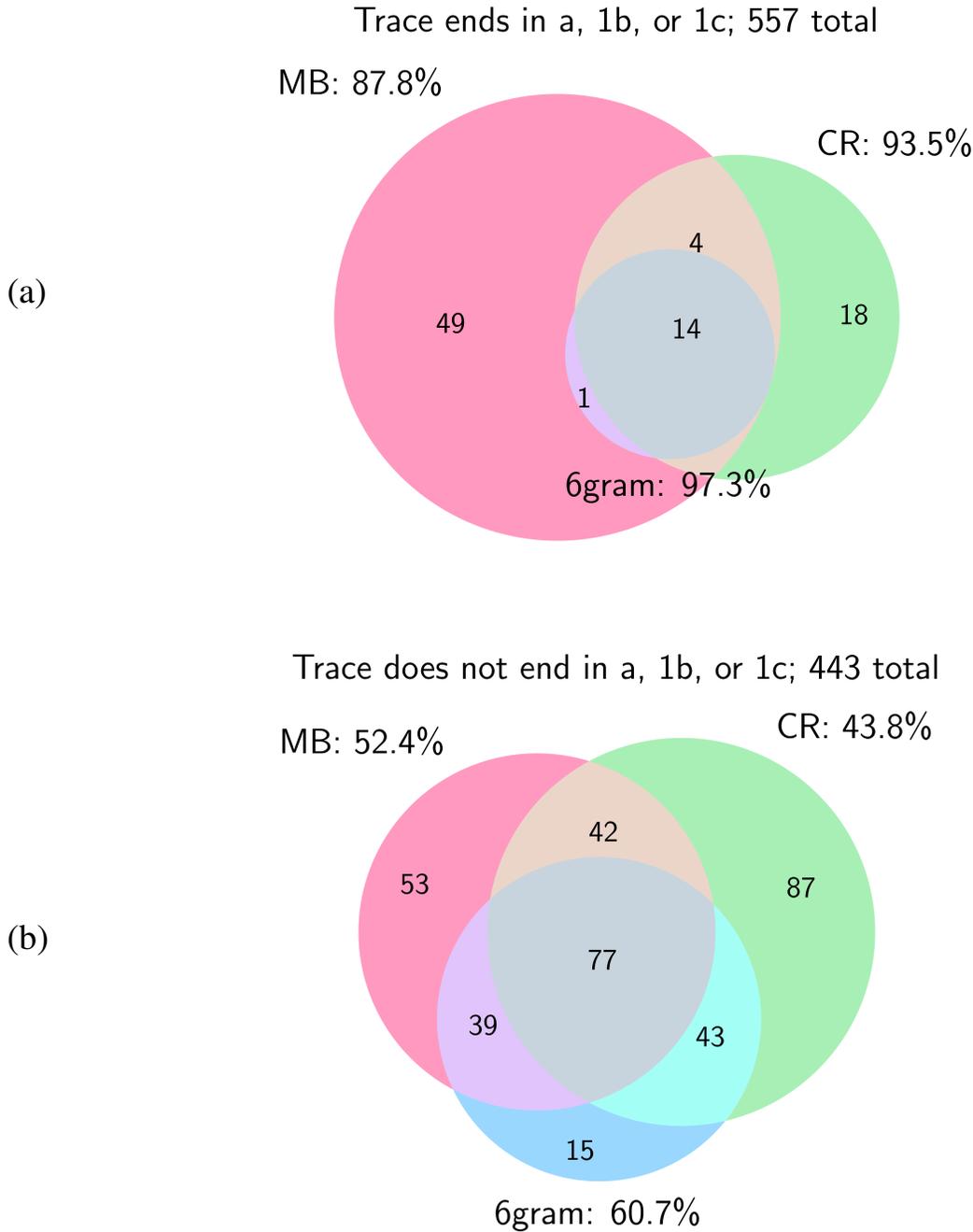


\include{content/D_0_venn}
\caption{
Results on Sum Modulo 3 DFA under trivial / nontrivial conditions. Percentages are accuracy numbers, and venn diagram is error counts.
(a) In this condition, CR and the 6-\textsc{Gram}$_T$ both get very high accuracies, with nearly all 6-\textsc{Gram}$_T$ also being CR errors. MB does relatively poorly.
(b) In this condition, models do significantly more poorly overall, with CR in particular performing worse than chance. Here, errors are more symmetric, with more 6-\textsc{Gram}$_T$ errors that are not accounted for by either or both model, indicating that a larger fraction of both successes and failures in this condition are down to random chance.
}
\label{fig:case-study-more}
\end{figure}

The model
also attempts to identify periodic patterns, but identifies period-2 patterns more than period-3 patterns, despite knowing that there are three ``rooms'' (states). At no point in any of the 60 reasoning traces analyzed does it realize that this is a version of the Sum Modulo 3 DFA\footnote{In fact, in none of the 1000 traces do the substrings ``sum'' or ``mod'' appear, except as a part of ``assuming''}, but it does show some glimmers of world modeling: in a few cases it correctly determines that there are two red rooms; however, this does not lead to further discoveries. It is not superior reasoning that leads to correct solutions, rather the correct examples are more likely to be ones that a 3-\textsc{Gram} model would infer correctly, i.e., those traces ending in \texttt{a}, \texttt{1b}, or \texttt{1c}, which occur cumulatively in $\frac{5}{9}$ of cases\footnote{On the $\sim$$\frac{5}{9}$ of examples following this pattern, CR achieves 93.5\%, to the 6-\textsc{Gram}$_T$'s 97.3\%, and on the remaining $\sim$$\frac49$, it achieves 43.8\%, to the 6-\textsc{Gram}$_T$'s 60.7\%. Detailed Venn diagrams on these conditions can be found in Figure~\ref{fig:case-study-more}.}.

Despite transformers' high computational capacity, without the ability to pattern match to existing problems, Claude uses an unsophisticated and ineffectual approach.

\clearpage
\section{Increased number of examples}

\begin{figure*}[htbp]
    \centering
    \includegraphics[width=0.9\textwidth]{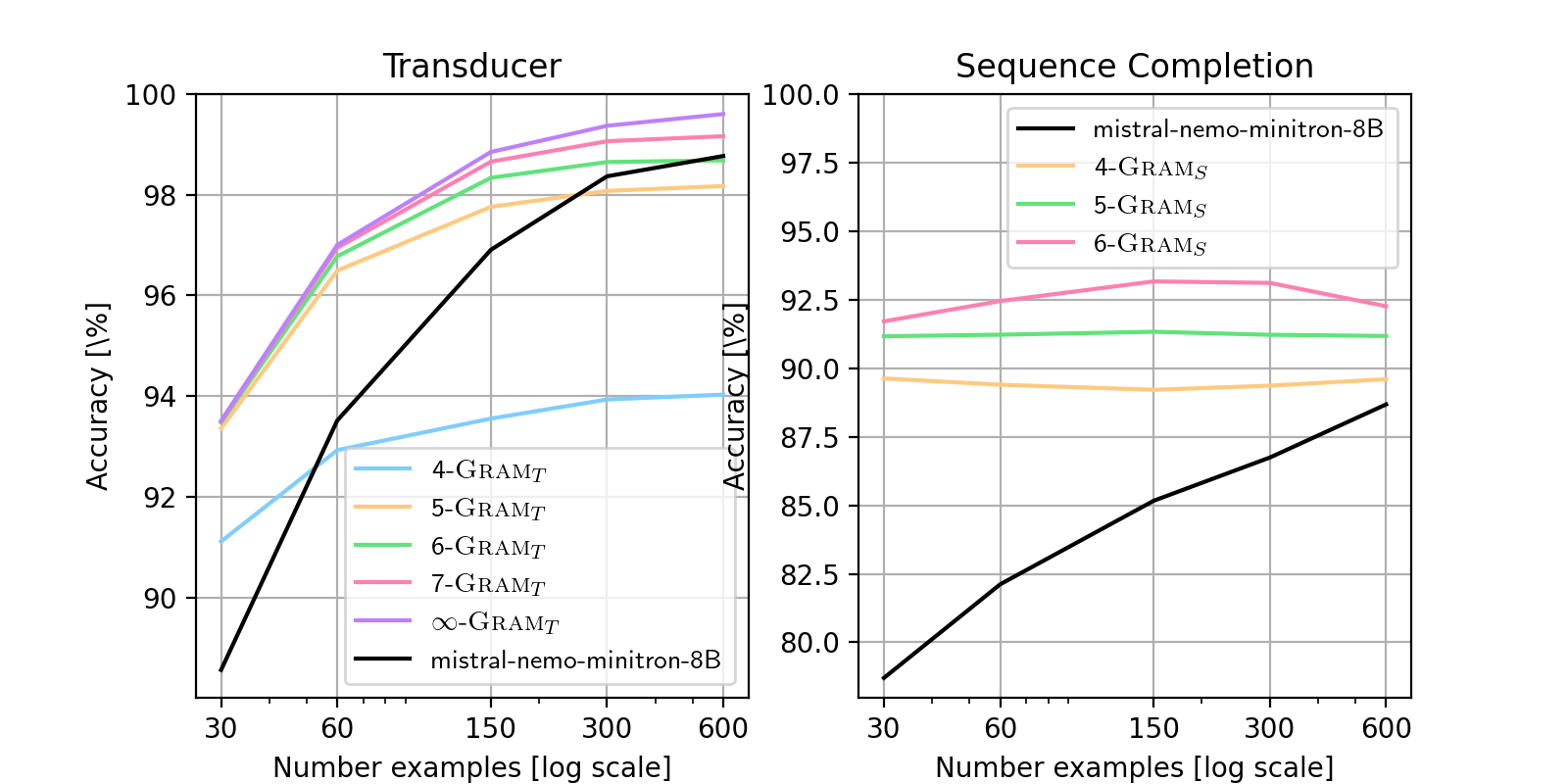}
    \caption{
        Accuracy by number of examples, as we vary the number from 30 to 600 (note the log scale). $\infty$-\textsc{Gram} is a model that finds the longest sequence that matches the ending of the sequence and copies the following token.
    }
    \label{fig:more-examples-per-problem}
\end{figure*}

We varied the number of examples parameter from 30 to 600, investigating specifically mistral-nemo-minitron-8B as it is a better performing model on the transducer task. We find that for the Sequence Completion task, the $n$-\textsc{Gram} models do not generally improve as the number of examples increases, while the LLM does; however, the LLM remains below 4-\textsc{Gram} performance at all points. In the Transducer experiment, both the n-\textsc{Gram} models and the LLM improve, with the LLM crossing 4-\textsc{Gram} and 5-Gram performance and eventually roughly matching 6-\textsc{Gram} performance. However, there is now an increased gap between 6-\textsc{Gram} and 7-\textsc{Gram} that did not exist at 30 (we did not include 7-\textsc{Gram} or above in the main table for this reason). Overall, it is possible that the larger number of instances of the ``correct n-\textsc{Gram}'' appearing (i.e., the suffix of the prompt followed by the correct answer) causes the model to be better at producing predictions. As the LLM at no point outperforms the best n-\textsc{Gram}, we do not believe that this alters our overall conclusions.

\clearpage
\section{Nonzero temperature}

\begin{table*}[htbp]
    \centering
    {\scriptsize
    {\renewcommand{\arraystretch}{1.25}
\begin{tabular}{llrrr}
\hline
\textbf{Model} & \textbf{Prompt} & \textbf{Zero Temp} & \textbf{Nonzero Temp} & \textbf{Difference} \\
\hline
\multicolumn{5}{l}{\textbf{Sequence Completion}} \\
\hline
mistral-nemo-minitron-8B & \textsc{Basic} & 78.70\% (77.49\% -- 79.79\%) & 77.67\% (76.51\% -- 78.76\%) & -1.04\% (-1.49\% -- -0.63\%) \\
\hline
claude-3.5 & \textsc{COT} & 84.00\% (79.33\% -- 88.44\%) & 84.22\% (79.56\% -- 89.00\%) & 0.22\% (-2.11\% -- 2.33\%) \\
\hline
claude-3.5 & \textsc{Red-Green} & 80.00\% (74.89\% -- 85.22\%) & 80.78\% (75.00\% -- 86.11\%) & 0.78\% (-2.11\% -- 3.44\%) \\
\hline
\multicolumn{5}{l}{\textbf{Transducer}} \\
\hline
mistral-nemo-minitron-8B & \textsc{Basic} & 88.56\% (88.05\% -- 89.08\%) & 88.17\% (87.64\% -- 88.68\%) & -0.39\% (-0.58\% -- -0.22\%) \\
\hline
claude-3.5 & \textsc{Basic} & 86.89\% (83.33\% -- 90.00\%) & 87.00\% (83.33\% -- 90.11\%) & 0.11\% (-0.67\% -- 0.89\%) \\
\hline
claude-3.5 & \textsc{More-Expl} & 87.11\% (83.88\% -- 90.22\%) & 86.89\% (83.22\% -- 90.11\%) & -0.22\% (-1.11\% -- 0.67\%) \\
\hline
claude-3.5 & \textsc{COT} & 76.44\% (72.89\% -- 79.89\%) & 78.11\% (74.66\% -- 81.33\%) & 1.67\% (-0.56\% -- 3.78\%) \\
\hline
claude-3.5 & \textsc{Red-Green} & 82.89\% (78.89\% -- 86.89\%) & 82.78\% (78.89\% -- 86.33\%) & -0.11\% (-2.00\% -- 1.78\%) \\
\hline
\end{tabular}
}

}

    \caption{Results varying temperature. Second column is a temperature of 0.1, third column is differences. In all columns, we annotate a 95\% confidence interval, using paired differences for the third column.}
    \label{tab:nonzero-temperature}
\end{table*}

In order to determine whether a small nonzero temperature might lead to better results, we investigated using a temperature of 0.1 for mistral-nemo-minitron-8B and claude-3.5 (these two chosen for the reasons described in Section \ref{app:case-study}, notably o3-mini/gpt-5 already are using nonzero temperatures). We find that a temperature of 0.1 does not significantly change performance, resulting in no significant change for any model/task/prompt combination. The largest improvement we observe is claude-3.5 on COT on the Transducer task, which gains 1.7\%, from 76.4\% to 78.1\% (still not enough to make it the best prompt for claude-3.5).

\clearpage
\section{Tokenization}

\begin{table*}[!htbp]
    \centering
    \begin{tabular}{|r|c|c|}
\hline
Model & \textsc{Basic}$_S$ & \textsc{Basic-Commas}$_S$\\
\hline
\bf qwen-2.5-coder-7B &\bf 79.5 (78.4--80.5)&\bf 60.7 (59.3--62.1)\\
\hline
qwen-2.5-coder-instruct-7B &79.5 (78.3--80.5)&55.5 (54.0--56.9)\\
\hline
qwen-2.5-coder-instruct-32B &79.2 (78.0--80.3)&55.2 (53.7--56.7)\\
\hline
mistral-nemo-minitron-8B &78.7 (77.5--79.8)&59.3 (57.9--60.8)\\
\hline
codestral-22B &78.0 (76.8--79.1)&59.0 (57.5--60.3)\\
\hline
qwen-2.5-32B &76.8 (75.5--78.0)&60.3 (58.9--61.6)\\
\hline
deepseek-coder-33b-instruct &76.7 (75.3--77.8)&54.9 (53.0--56.8)\\
\hline
mistral-nemo-base-12B &75.5 (74.3--76.6)&60.6 (59.1--62.2)\\
\hline
llama3.1-8B-Instruct &75.3 (74.0--76.6)&56.3 (54.4--58.1)\\
\hline
llama3.1-8B &75.2 (73.8--76.3)&61.1 (59.8--62.5)\\
\hline
llama3-8B &73.8 (72.4--75.1)&61.5 (60.2--62.9)\\
\hline
starcoder2-15b &73.5 (72.0--74.7)&58.2 (56.7--59.8)\\
\hline
qwen-2.5-7B &73.5 (72.1--74.8)&57.0 (55.5--58.5)\\
\hline
gemma-7b &72.6 (71.3--73.7)&54.0 (51.9--56.0)\\
\hline
gpt-4o-mini &72.4 (68.1--76.3)&64.1 (59.5--68.3)\\
\hline
mistral-nemo-instruct-12B &72.2 (70.9--73.4)&58.2 (56.4--59.8)\\
\hline
gpt-4o &72.1 (65.9--78.2)&66.8 (58.5--74.8)\\
\hline
llama3.1-70B &71.8 (70.4--73.1)&57.7 (56.1--59.2)\\
\hline
llama3-70B &71.4 (70.0--72.7)&56.4 (54.7--58.0)\\
\hline
falcon-7b &69.0 (67.6--70.2)&56.1 (54.5--57.6)\\
\hline
gpt-3.5-instruct &67.3 (63.1--71.5)&52.3 (46.5--57.9)\\
\hline
o3-mini &N/A&N/A\\
\hline
claude-3.5 &N/A&N/A\\
\hline
gpt-3.5-chat &N/A&N/A\\
\hline
\end{tabular}

    \caption{Results on Sequence Completion Task. We compare \textsc{Basic}$_S$ to the comma-variant \textsc{Basic-Commas}$_S$.}
    \label{tab:results-sequence-completion-commas}
\end{table*}

\subsection{Sequence Completion task prompt with Commas} \label{app:sequence-completion-commas}

To avoid tokenization differences with models, we also investigate a version of our Sequence Completion prompt that uses spaces and commas between the elements of the sequence. Unfortunately, results using this prompt were uniformly worse than results on the prompt without spaces and commas. Table~\ref{tab:results-sequence-completion-commas} shows the results on a variety of models. All are worse with commas than without.

\subsection{Directly confirming models can read sequences of letters} \label{app:results-tokenization-test-regexp}

To fully exclude the possibility that models are unable to read sequences of letters, we perform the following experiment: we take the regular expression \texttt{ab(abc)+} and directly provide it to the model, then ask the model to test it on a string (one string provided per query), using the prompt \texttt{I will give you a string. Tell me whether it matches the following regular expression: '\^ab(abc)+\$' (without quotes). Just answer YES or NO.} on one line, followed by a string on the next. We sample 100 random strings of length 2-17, via the following procedure (1) sample a random valid string uniformly (2) with 50\% probability, randomly mutate one of the elements of the string to a different character. All strings are thus either correct or near-correct. We also add \texttt{Answer (YES or NO): } on a third line for non-chat models to encourage them to provide a response rather than another string.

\begin{table*}[!htbp]
    \centering
    \begin{tabular}{l r r}
\toprule
Model & Accuracy & Non-response \\
\midrule
llama3-8B & 66\% &  \\
llama3-70B & 80\% & 26\% \\
llama3.1-8B-Instruct & 83\% &  \\
llama3.1-8B & 89\% &  \\
llama3.1-70B & 86\% & 36\% \\
qwen-2.5-7B & 62\% &  \\
qwen-2.5-32B & 100\% &  \\
mistral-nemo-instruct-12B & 99\% &  \\
gemma-7b & 67\% & 15\% \\
starcoder2-15b & 62\% &  \\
codestral-22B & 95\% &  \\
qwen-2.5-coder-7B & 100\% &  \\
qwen-2.5-coder-instruct-7B & 56\% &  \\
qwen-2.5-coder-instruct-32B & 96\% &  \\
gpt-3.5-instruct & 67\% &  \\
gpt-3.5-chat & 82\% &  \\
gpt-4o-mini & 100\% &  \\
gpt-4o & 100\% &  \\
claude-3.5 & 100\% &  \\
o3-mini & 100\% &  \\
gpt-5 & 100\% &  \\
\bottomrule
\end{tabular}
    \caption{Results on Regex task. As in the rest of this paper, accuracies are computed ignoring non-response. Models mistral-nemo-minitron-8B, mistral-nemo-base-12B, falcon-7b, deepseek-coder-33b-instruct have non-response rates over 98\%.}
    \label{tab:results-tokenization-test-regexp}
\end{table*}

Results for this are presented in Table~\ref{tab:results-tokenization-test-regexp}. While not all models perform well at this task, performances follow roughly what one might expect from standard benchmarks (rather than the Sequence Completion results in the paper). Frontier proprietary models perform extremely well, larger open weight models tend to perform well, and smaller open weight models are more hit-and-miss (some performing well, some poorly).

The proprietary models newer than the GPT-3 series get 100\% accuracy on this task, while performance is lower for the open weight models. There is no obvious relationship between models that perform well at this particular task and models that perform well at Sequence Completion, indicating that to whatever degree models are performing poorly at this test task, it is not because they are unable to process the string. Notably, o3-mini, gpt-4o, and gpt-4o-mini all perform fairly poorly at the Sequence Completion task, placing below median, but all achieve 100\% on this task.

Open weight models are more mixed, with larger ones tending to do nearly as well as newer proprietary models (notable exception being llama3-70B), but smaller ones occasionally performing well and occasionally performing poorly.

\clearpage

\section{Model non-answers}

\begin{table*}[htbp]
    \centering
    {\scriptsize
{\renewcommand{\arraystretch}{1.25}\begin{tabular}{l|cc|ccc}
\hline
\bf Model & \bf \textsc{Basic} & \bf \textsc{Basic-COT} & \bf \textsc{More-Expl} & \bf \textsc{DFA-COT} & \bf \textsc{Red-Green} \\
\hline
\multicolumn{6}{l}{ \bf Sequence Completion} \\
\hline
llama3-8B & 0.0 (0.0--0.0) & -- & -- & -- & -- \\
\hline
llama3-70B & 0.0 (0.0--0.0) & -- & -- & -- & -- \\
\hline
llama3.1-8B-Instruct & 0.0 (0.0--0.0) & -- & -- & -- & -- \\
\hline
llama3.1-8B & 0.0 (0.0--0.0) & -- & -- & -- & -- \\
\hline
llama3.1-70B & 0.0 (0.0--0.0) & -- & -- & -- & -- \\
\hline
mistral-nemo-minitron-8B & 0.0 (0.0--0.0) & -- & -- & -- & -- \\
\hline
mistral-nemo-base-12B & 0.0 (0.0--0.0) & -- & -- & -- & -- \\
\hline
mistral-nemo-instruct-12B & 0.0 (0.0--0.0) & -- & -- & -- & -- \\
\hline
gemma-7b & 0.0 (0.0--0.0) & -- & -- & -- & -- \\
\hline
falcon-7b & 0.0 (0.0--0.0) & -- & -- & -- & -- \\
\hline
starcoder2-15b & 0.0 (0.0--0.0) & -- & -- & -- & -- \\
\hline
codestral-22B & 0.0 (0.0--0.0) & -- & -- & -- & -- \\
\hline
deepseek-coder-33b-instruct & 0.0 (0.0--0.0) & -- & -- & -- & -- \\
\hline
qwen-2.5-coder-instruct-7B & 0.0 (0.0--0.0) & -- & -- & -- & -- \\
\hline
qwen-2.5-7B & 0.0 (0.0--0.0) & -- & -- & -- & -- \\
\hline
qwen-2.5-32B & 0.0 (0.0--0.0) & -- & -- & -- & -- \\
\hline
qwen-2.5-coder-instruct-32B & 0.0 (0.0--0.0) & -- & -- & -- & -- \\
\hline
qwen-2.5-coder-7B & 0.0 (0.0--0.0) & -- & -- & -- & -- \\
\hline
gpt-3.5-instruct & 2.5 (1.9--3.0) & -- & -- & -- & -- \\
\hline
gpt-3.5-chat & 99.9 (99.7--100.0) & -- & -- & -- & -- \\
\hline
gpt-4o-mini & 0.0 (0.0--0.0) & 0.3 (0.2--0.5) & 0.0 (0.0--0.0) & 1.0 (0.6--1.4) & 0.2 (0.1--0.4) \\
\hline
gpt-4o & 4.4 (3.3--5.7) & 1.9 (1.1--2.8) & 100.0 (100.0--100.0) & 5.0 (3.8--6.2) & 8.4 (6.6--10.2) \\
\hline
claude-3.5 & 99.7 (99.2--100.0) & 0.1 (0.0--0.3) & 97.8 (96.9--98.6) & 0.0 (0.0--0.0) & 0.0 (0.0--0.0) \\
\hline
o3-mini & 80.2 (78.0--82.3) & 3.8 (2.6--5.2) & 91.6 (89.9--93.2) & 5.7 (4.1--7.2) & 0.4 (0.0--1.0) \\
\hline
gpt-5 & 0.0 (0.0--0.0) & 0.0 (0.0--0.0) & 8.2 (4.7--12.1) & 5.2 (3.1--7.4) & 0.4 (0.1--0.9) \\
\hline
\multicolumn{6}{l}{ \bf Transducer} \\
\hline
llama3-8B & 0.0 (0.0--0.0) & -- & -- & -- & -- \\
\hline
llama3-70B & 0.0 (0.0--0.0) & -- & -- & -- & -- \\
\hline
llama3.1-8B-Instruct & 0.0 (0.0--0.0) & -- & -- & -- & -- \\
\hline
llama3.1-70B & 0.0 (0.0--0.0) & -- & -- & -- & -- \\
\hline
llama3.1-8B & 0.0 (0.0--0.0) & -- & -- & -- & -- \\
\hline
starcoder2-15b & 0.0 (0.0--0.0) & -- & -- & -- & -- \\
\hline
codestral-22B & 0.0 (0.0--0.0) & -- & -- & -- & -- \\
\hline
deepseek-coder-33b-instruct & 0.0 (0.0--0.0) & -- & -- & -- & -- \\
\hline
qwen-2.5-coder-7B & 0.0 (0.0--0.0) & -- & -- & -- & -- \\
\hline
qwen-2.5-coder-instruct-7B & 0.0 (0.0--0.0) & -- & -- & -- & -- \\
\hline
qwen-2.5-7B & 0.0 (0.0--0.0) & -- & -- & -- & -- \\
\hline
qwen-2.5-32B & 0.0 (0.0--0.0) & -- & -- & -- & -- \\
\hline
qwen-2.5-coder-instruct-32B & 0.0 (0.0--0.0) & -- & -- & -- & -- \\
\hline
mistral-nemo-minitron-8B & 0.0 (0.0--0.0) & -- & -- & -- & -- \\
\hline
mistral-nemo-base-12B & 0.0 (0.0--0.0) & -- & -- & -- & -- \\
\hline
mistral-nemo-instruct-12B & 0.0 (0.0--0.0) & -- & -- & -- & -- \\
\hline
gemma-7b & 0.0 (0.0--0.0) & -- & -- & -- & -- \\
\hline
falcon-7b & 0.0 (0.0--0.0) & -- & -- & -- & -- \\
\hline
gpt-3.5-instruct & 0.0 (0.0--0.1) & -- & -- & -- & -- \\
\hline
gpt-3.5-chat & 0.1 (0.0--0.3) & -- & -- & -- & -- \\
\hline
gpt-4o-mini & 1.8 (1.3--2.3) & 0.0 (0.0--0.0) & 5.8 (4.8--6.9) & 0.0 (0.0--0.0) & 0.7 (0.4--1.0) \\
\hline
gpt-4o & 0.0 (0.0--0.0) & 0.0 (0.0--0.0) & 0.0 (0.0--0.0) & 0.0 (0.0--0.0) & 0.0 (0.0--0.0) \\
\hline
claude-3.5 & 0.0 (0.0--0.0) & 0.0 (0.0--0.0) & 0.0 (0.0--0.0) & 0.0 (0.0--0.0) & 0.0 (0.0--0.0) \\
\hline
o3-mini & 0.1 (0.0--0.3) & 0.0 (0.0--0.0) & 0.4 (0.1--0.9) & 0.0 (0.0--0.0) & 0.0 (0.0--0.1) \\
\hline
gpt-5 & 0.0 (0.0--0.0) & 0.0 (0.0--0.0) & 0.0 (0.0--0.0) & 0.0 (0.0--0.0) & 0.7 (0.1--1.4) \\
\hline
\end{tabular}
}

}

    \caption{Model non-answers, as a percentage of all prompt responses. A non-response is not included in accuracy computations for Table~\ref{tab:results-main} or Table~\ref{tab:results-multiprompt}, but whenever it rises above 25\%, N/A is placed in those tables.}
    \label{tab:results-null}
\end{table*}

Table~\ref{tab:results-null} depicts the percentage of model non-answers by model and prompt. In general, this distribution is highly bimodal, with values always being either below 9\% or above 97\%.

The only prompt-vs-prompt orderings that are changed by scoring non-answers as 0 are that, on Sequence Completion, \textsc{Basic}$_S$ rises above \textsc{Red-Green}$_S$ for gpt-4o, making it the best prompt; and that on Transducer, \textsc{Red-Green}$_T$ for gpt-4o-mini rises above \textsc{More-Expl}$_T$ (though still behind \textsc{Basic}$_T$. The qualitative conclusions about next token prediction vs chain of thought results remain the same.

The only change to relative model ordering is that on Sequence Completion, gpt-4o drops 8 ranks, from 17th place to 25nd place, being passed by several open weight models, gpt-4o-mini, and o3-mini. No change occurs on the transducer results. Qualiative conclusions about model ordering remain the same.

\clearpage
\section{Significance} \label{app:significance}

\begin{figure*}[ht]
    \centering
    \includegraphics[width=0.9\textwidth]{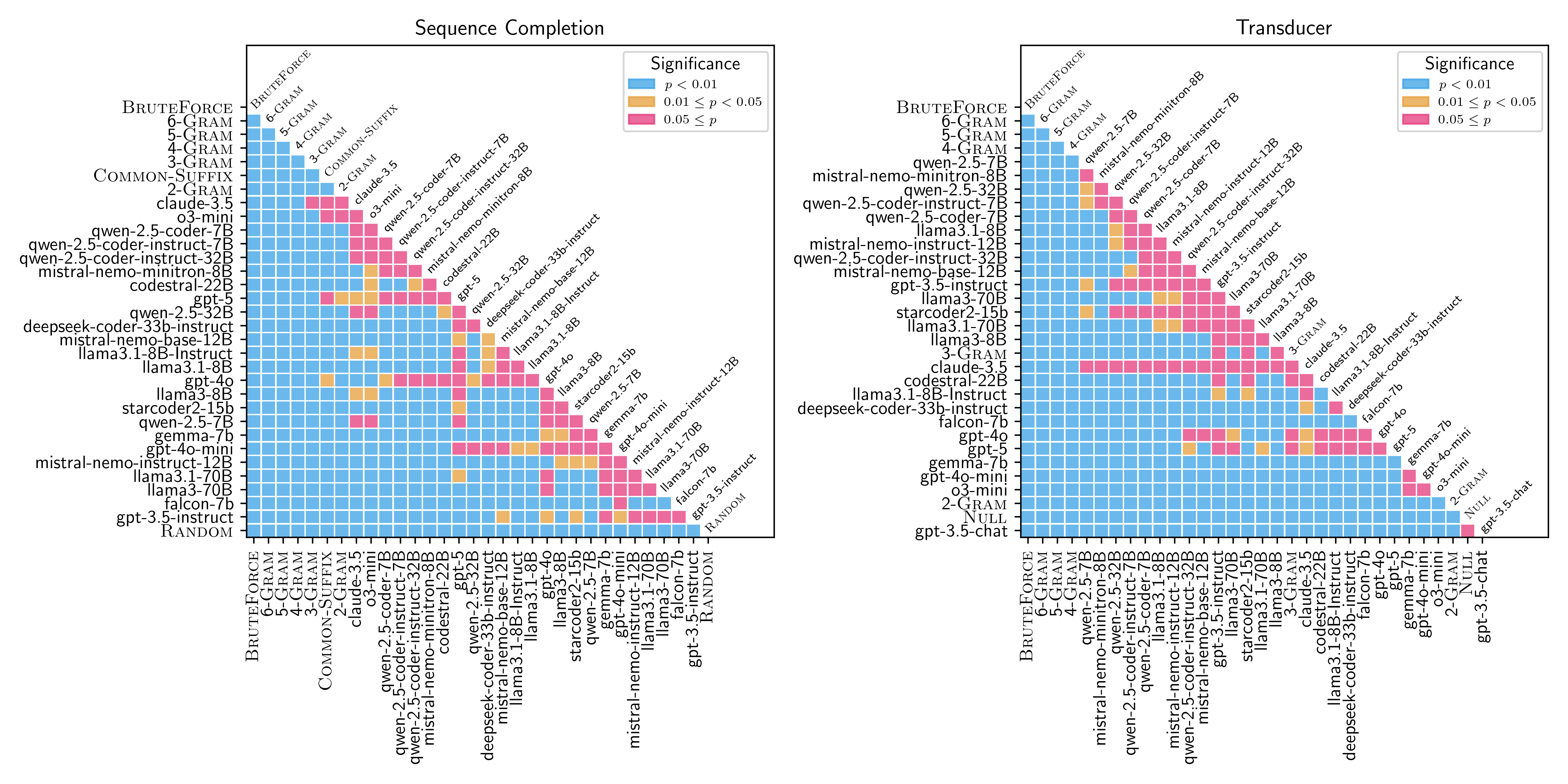}
    \caption{
        Significance of comparisons between rows of Table~\ref{tab:results-main}. Results in blue and orange are significant, results in pink are not.
    }
    \label{fig:significance}
\end{figure*}

Figure~\ref{fig:significance} shows which comparisons between rows of Table~\ref{tab:results-main} are significant. Significance computations are performed by running a 2-tailed bootstrap significance test on paired (by DFAs) differences.

\clearpage

\section{Prompt Listings} \label{app:prompt-listings}

\begin{table*}[htpb]
    \centering
    \tiny
    \begin{tabular}{|r|l|l|}
\hline
\bf Prompt & \bf $T$ & \bf $S$\\
\hline
\textsc{Basic}&\begin{minipage}{2.5in}\begin{lstlisting}
You are a sequence completion model. Output the next element of the sequence, and nothing else.

<TRANSDUCER PREFIX>, 
\end{lstlisting}\end{minipage}&\begin{minipage}{2.5in}\begin{lstlisting}
The following strings come from an alien language that follows a simple grammar. Infer the alien grammar using the example strings. Then, add a suffix to the final string using between 1 and 5 characters such that the full string follows the grammar. Output only the necessary suffix to complete the final string, and nothing else.

<EXAMPLES>
<PREFIX>
\end{lstlisting}\end{minipage}\\
\hline\textsc{Basic-COT}&\begin{minipage}{2.5in}\begin{lstlisting}
You are a sequence completion model. Reason step by step, and then output the next output integer using <answer> tags, like <answer>0</answer>.

Input sequence: <TRANSDUCER PREFIX>
Output sequence: 
\end{lstlisting}\end{minipage}&\begin{minipage}{2.5in}\begin{lstlisting}
The following strings come from an alien language that follows a simple grammar. Infer the alien grammar using the example strings. Then, add a suffix to the final string using between 1 and 5 characters such that the full string follows the grammar. Reason step by step, and then output the next necessary suffix for this final string, <answer> tags, like <answer>ab</answer>.

<EXAMPLES>
<PREFIX>
\end{lstlisting}\end{minipage}\\
\hline\textsc{More-Expl}&\begin{minipage}{2.5in}\begin{lstlisting}
You are a sequence completion model. The following sequence is generated from an unknown but consistent grammar. Identify the patterns within the sequence to determine its next element. Output the next element of the sequence, and nothing else.

<TRANSDUCER PREFIX>, 
\end{lstlisting}\end{minipage}&\begin{minipage}{2.5in}\begin{lstlisting}
I have a 3-state DFA model that outputs either 0 or 1 after each element I input. 1 indicates that the input string thus far results in a "valid" state, and 0 indicates that it does not. I collect a set of valid strings using this DFA, listed below. Infer the underlying DFA model using these strings and complete the final string, using up to n characters, such that it is also a valid string. Output only the necessary suffix to complete the final string, and nothing else.
<EXAMPLES>
<PREFIX>
\end{lstlisting}\end{minipage}\\
\hline\textsc{DFA-COT}&\begin{minipage}{2.5in}\begin{lstlisting}
A DFA is a finite-state machine that accepts or rejects a given string of symbols, by running through a n-state sequence uniquely determined by the string.

I have a 3-state DFA model that outputs either 0 or 1 after each element I input. 1 indicates that the input string thus far results in a "valid" state, and 0 indicates that it does not. I collect the inputs and outputs into an input sequence and an output sequence. Infer the underlying DFA model to predict the next integer in the output sequence. Reason step by step, and then output the next output integer using <answer> tags, like <answer>0</answer>.

Input sequence: <TRANSDUCER PREFIX>
Output sequence: 
\end{lstlisting}\end{minipage}&\begin{minipage}{2.5in}\begin{lstlisting}
I have a 3-state DFA model that outputs either 0 or 1 after each element I input. 1 indicates that the input string thus far results in a "valid" state, and 0 indicates that it does not. I collect a set of valid strings using this DFA, listed below. Infer the underlying DFA model using these strings and complete the final string, using up to n characters, such that it is also a valid string. Reason step by step, and then output the next necessary suffix for this final string, <answer> tags, like <answer>ab</answer>.

Given these valid strings:
<EXAMPLES>

Complete the following string:
<PREFIX>
\end{lstlisting}\end{minipage}\\
\hline\textsc{Red-Green}&\begin{minipage}{2.5in}\begin{lstlisting}
```
You are in a house of rooms and portals. There are 3 rooms in the house, and each room has 3 unique portals labeled A, B, and C. Each portal teleports you to one room of the house (and sometimes the destination is the room the portal is in). Every portal in a given room always behaves the same way.

In this house, each of the rooms look exactly the same, except some of the rooms have red walls and some have green walls. However, there are *three* rooms in total, so you cannot determine which room you are in by color alone, and two rooms of the same color may have portals that behave differently.  As you move through the house, at each time step you write down what portal you take and the color of the room you arrive (or stay) in. Based on your notes, predict what color room you will end up in after the last step.

Tag your final answer like <answer>color</answer>.

You walk through a portal labeled "<TRANSDUCER PREFIX>" and end up in a red room.
```
\end{lstlisting}\end{minipage}&\begin{minipage}{2.5in}\begin{lstlisting}
You are outside a house of rooms and portals. There are 3 rooms in the house, and each room has 3 unique portals labeled a, b, and c. Each portal teleports you to one room of the house (and sometimes the destination is the room the portal is in). Every portal in a given room always behaves the same way.

In this house, each of the rooms look exactly the same, except some of the rooms have red walls and some have green walls. However, there are *3* rooms in total, so you cannot determine which room you are in by color alone, and two rooms of the same color may have portals that behave differently. You've been into this house many times before. Each time, as you move through the house, you write down what series of portals you take and the color of the room you end up in. You have a collection of paths you've taken where you've ended up in a room with green walls, listed below. Given the final incomplete path at the bottom, write a series of up to 5 remaining steps that will cause you to end up in a room with green walls again.

Tag your final answer like <answer>ab</answer>.

Given these paths that end in a room with green walls:
<EXAMPLES>

Complete the following path:
<PREFIX>
\end{lstlisting}\end{minipage}\\
\hline\end{tabular}

    \caption{Shortened summary of each prompt}
    \label{tab:prompt-preambles}
\end{table*}

\subsection{Summaries}

Table~\ref{tab:prompt-preambles} contains summaries of each prompt.

\subsection{Full example listings}

\subsubsection{\textsc{Basic}\texorpdfstring{$_T$}{T}}
\begin{lstlisting}
You are a sequence completion model. Output the next element of the sequence, and nothing else.

a, 1, b, 1, a, 1, b, 1, b, 1, c, 0, a, 1, c, 1, a, 1, a, 1, a, 1, c, 1, b, 1, c, 0, c, 1, a, 1, b, 1, b, 1, b, 1, b, 1, a, 1, b, 1, a, 1, a, 1, b, 1, c, 0, a, 1, c, 1, a, 1, b, 
\end{lstlisting}
\subsubsection{\textsc{Basic-COT}\texorpdfstring{$_T$}{T}}
\begin{lstlisting}
You are a sequence completion model. Reason step by step, and then output the next output integer using <answer> tags, like <answer>0</answer>.

Input sequence: a, b, a, b, b, c, a, c, a, a, a, c, b, c, c, a, b, b, b, b, a, b, a, a, b, c, a, c, a, b
Output sequence: 1, 1, 1, 1, 1, 0, 1, 1, 1, 1, 1, 1, 1, 0, 1, 1, 1, 1, 1, 1, 1, 1, 1, 1, 1, 0, 1, 1, 1, 
\end{lstlisting}
\subsubsection{\textsc{More-Expl}\texorpdfstring{$_T$}{T}}
\begin{lstlisting}
You are a sequence completion model. The following sequence is generated from an unknown but consistent grammar. Identify the patterns within the sequence to determine its next element. Output the next element of the sequence, and nothing else.

a, 1, b, 1, a, 1, b, 1, b, 1, c, 0, a, 1, c, 1, a, 1, a, 1, a, 1, c, 1, b, 1, c, 0, c, 1, a, 1, b, 1, b, 1, b, 1, b, 1, a, 1, b, 1, a, 1, a, 1, b, 1, c, 0, a, 1, c, 1, a, 1, b, 
\end{lstlisting}
\subsubsection{\textsc{DFA-COT}\texorpdfstring{$_T$}{T}}
\begin{lstlisting}
A DFA is a finite-state machine that accepts or rejects a given string of symbols, by running through a n-state sequence uniquely determined by the string.

I have a 3-state DFA model that outputs either 0 or 1 after each element I input. 1 indicates that the input string thus far results in a "valid" state, and 0 indicates that it does not. I collect the inputs and outputs into an input sequence and an output sequence. Infer the underlying DFA model to predict the next integer in the output sequence. Reason step by step, and then output the next output integer using <answer> tags, like <answer>0</answer>.

Input sequence: a, b, a, b, b, c, a, c, a, a, a, c, b, c, c, a, b, b, b, b, a, b, a, a, b, c, a, c, a, b
Output sequence: 1, 1, 1, 1, 1, 0, 1, 1, 1, 1, 1, 1, 1, 0, 1, 1, 1, 1, 1, 1, 1, 1, 1, 1, 1, 0, 1, 1, 1, 
\end{lstlisting}
\subsubsection{\textsc{Red-Green}\texorpdfstring{$_T$}{T}}
\begin{lstlisting}
```
You are in a house of rooms and portals. There are 3 rooms in the house, and each room has 3 unique portals labeled A, B, and C. Each portal teleports you to one room of the house (and sometimes the destination is the room the portal is in). Every portal in a given room always behaves the same way.

In this house, each of the rooms look exactly the same, except some of the rooms have red walls and some have green walls. However, there are *three* rooms in total, so you cannot determine which room you are in by color alone, and two rooms of the same color may have portals that behave differently.  As you move through the house, at each time step you write down what portal you take and the color of the room you arrive (or stay) in. Based on your notes, predict what color room you will end up in after the last step.

Tag your final answer like <answer>color</answer>.

You walk through a portal labeled "A" and end up in a green room.
Then, you walk through a portal labeled "B" and end up in a green room.
Then, you walk through a portal labeled "A" and end up in a green room.
Then, you walk through a portal labeled "B" and end up in a green room.
Then, you walk through a portal labeled "B" and end up in a green room.
Then, you walk through a portal labeled "C" and end up in a red room.
Then, you walk through a portal labeled "A" and end up in a green room.
Then, you walk through a portal labeled "C" and end up in a green room.
Then, you walk through a portal labeled "A" and end up in a green room.
Then, you walk through a portal labeled "A" and end up in a green room.
Then, you walk through a portal labeled "A" and end up in a green room.
Then, you walk through a portal labeled "C" and end up in a green room.
Then, you walk through a portal labeled "B" and end up in a green room.
Then, you walk through a portal labeled "C" and end up in a red room.
Then, you walk through a portal labeled "C" and end up in a green room.
Then, you walk through a portal labeled "A" and end up in a green room.
Then, you walk through a portal labeled "B" and end up in a green room.
Then, you walk through a portal labeled "B" and end up in a green room.
Then, you walk through a portal labeled "B" and end up in a green room.
Then, you walk through a portal labeled "B" and end up in a green room.
Then, you walk through a portal labeled "A" and end up in a green room.
Then, you walk through a portal labeled "B" and end up in a green room.
Then, you walk through a portal labeled "A" and end up in a green room.
Then, you walk through a portal labeled "A" and end up in a green room.
Then, you walk through a portal labeled "B" and end up in a green room.
Then, you walk through a portal labeled "C" and end up in a red room.
Then, you walk through a portal labeled "A" and end up in a green room.
Then, you walk through a portal labeled "C" and end up in a green room.
Then, you walk through a portal labeled "A" and end up in a green room.
Then, you walk through a portal labeled "B" and end up in a ...
```
\end{lstlisting}
\subsubsection{\textsc{Basic}\texorpdfstring{$_S$}{S}}
\begin{lstlisting}
The following strings come from an alien language that follows a simple grammar. Infer the alien grammar using the example strings. Then, add a suffix to the final string using between 1 and 5 characters such that the full string follows the grammar. Output only the necessary suffix to complete the final string, and nothing else.

cbcbabbcca
abcaaacbaa
aabccbabbb
bbbccbbbca
aababaccba
aaaacbacac
baacbccbaa
cbbaacabcc
baabaacaab
bbbbbcacab
acaabcbbba
acaacbccac
cacbabcbba
abcbcbcbcc
ccaccccaba
bcbcabbcca
baabacabca
caababacac
bacacaccaa
bcacbbbbca
bcbbbcaccc
ccabbcccbb
bccbcabbca
baacbabcbc
ccacabccab
caacbcaaab
cacbaaccac
aaccbcaabb
abacabcaab
bacbcbcaca
caacb
\end{lstlisting}
\subsubsection{\textsc{Basic-COT}\texorpdfstring{$_S$}{S}}
\begin{lstlisting}
The following strings come from an alien language that follows a simple grammar. Infer the alien grammar using the example strings. Then, add a suffix to the final string using between 1 and 5 characters such that the full string follows the grammar. Reason step by step, and then output the next necessary suffix for this final string, <answer> tags, like <answer>ab</answer>.

cbcbabbcca
abcaaacbaa
aabccbabbb
bbbccbbbca
aababaccba
aaaacbacac
baacbccbaa
cbbaacabcc
baabaacaab
bbbbbcacab
acaabcbbba
acaacbccac
cacbabcbba
abcbcbcbcc
ccaccccaba
bcbcabbcca
baabacabca
caababacac
bacacaccaa
bcacbbbbca
bcbbbcaccc
ccabbcccbb
bccbcabbca
baacbabcbc
ccacabccab
caacbcaaab
cacbaaccac
aaccbcaabb
abacabcaab
bacbcbcaca
caacb
\end{lstlisting}
\subsubsection{\textsc{Basic-Commas}\texorpdfstring{$_S$}{S}}
\begin{lstlisting}
The following strings come from an alien language that follows a simple grammar. Infer the alien grammar using the example strings. Then, add a suffix to the final string using between 1 and 5 characters such that the full string follows the grammar. Output only the necessary suffix to complete the final string, and nothing else.

c, b, c, b, a, b, b, c, c, a
a, b, c, a, a, a, c, b, a, a
a, a, b, c, c, b, a, b, b, b
b, b, b, c, c, b, b, b, c, a
a, a, b, a, b, a, c, c, b, a
a, a, a, a, c, b, a, c, a, c
b, a, a, c, b, c, c, b, a, a
c, b, b, a, a, c, a, b, c, c
b, a, a, b, a, a, c, a, a, b
b, b, b, b, b, c, a, c, a, b
a, c, a, a, b, c, b, b, b, a
a, c, a, a, c, b, c, c, a, c
c, a, c, b, a, b, c, b, b, a
a, b, c, b, c, b, c, b, c, c
c, c, a, c, c, c, c, a, b, a
b, c, b, c, a, b, b, c, c, a
b, a, a, b, a, c, a, b, c, a
c, a, a, b, a, b, a, c, a, c
b, a, c, a, c, a, c, c, a, a
b, c, a, c, b, b, b, b, c, a
b, c, b, b, b, c, a, c, c, c
c, c, a, b, b, c, c, c, b, b
b, c, c, b, c, a, b, b, c, a
b, a, a, c, b, a, b, c, b, c
c, c, a, c, a, b, c, c, a, b
c, a, a, c, b, c, a, a, a, b
c, a, c, b, a, a, c, c, a, c
a, a, c, c, b, c, a, a, b, b
a, b, a, c, a, b, c, a, a, b
b, a, c, b, c, b, c, a, c, a
c, a, a, c, b,
\end{lstlisting}
\subsubsection{\textsc{More-Expl}\texorpdfstring{$_S$}{S}}
\begin{lstlisting}
I have a 3-state DFA model that outputs either 0 or 1 after each element I input. 1 indicates that the input string thus far results in a "valid" state, and 0 indicates that it does not. I collect a set of valid strings using this DFA, listed below. Infer the underlying DFA model using these strings and complete the final string, using up to n characters, such that it is also a valid string. Output only the necessary suffix to complete the final string, and nothing else.
cbcbabbcca
abcaaacbaa
aabccbabbb
bbbccbbbca
aababaccba
aaaacbacac
baacbccbaa
cbbaacabcc
baabaacaab
bbbbbcacab
acaabcbbba
acaacbccac
cacbabcbba
abcbcbcbcc
ccaccccaba
bcbcabbcca
baabacabca
caababacac
bacacaccaa
bcacbbbbca
bcbbbcaccc
ccabbcccbb
bccbcabbca
baacbabcbc
ccacabccab
caacbcaaab
cacbaaccac
aaccbcaabb
abacabcaab
bacbcbcaca
caacb
\end{lstlisting}
\subsubsection{\textsc{DFA-COT}\texorpdfstring{$_S$}{S}}
\begin{lstlisting}
I have a 3-state DFA model that outputs either 0 or 1 after each element I input. 1 indicates that the input string thus far results in a "valid" state, and 0 indicates that it does not. I collect a set of valid strings using this DFA, listed below. Infer the underlying DFA model using these strings and complete the final string, using up to n characters, such that it is also a valid string. Reason step by step, and then output the next necessary suffix for this final string, <answer> tags, like <answer>ab</answer>.

Given these valid strings:
cbcbabbcca
abcaaacbaa
aabccbabbb
bbbccbbbca
aababaccba
aaaacbacac
baacbccbaa
cbbaacabcc
baabaacaab
bbbbbcacab
acaabcbbba
acaacbccac
cacbabcbba
abcbcbcbcc
ccaccccaba
bcbcabbcca
baabacabca
caababacac
bacacaccaa
bcacbbbbca
bcbbbcaccc
ccabbcccbb
bccbcabbca
baacbabcbc
ccacabccab
caacbcaaab
cacbaaccac
aaccbcaabb
abacabcaab
bacbcbcaca

Complete the following string:
caacb
\end{lstlisting}
\subsubsection{\textsc{Red-Green}\texorpdfstring{$_S$}{S}}
\begin{lstlisting}
You are outside a house of rooms and portals. There are 3 rooms in the house, and each room has 3 unique portals labeled a, b, and c. Each portal teleports you to one room of the house (and sometimes the destination is the room the portal is in). Every portal in a given room always behaves the same way.

In this house, each of the rooms look exactly the same, except some of the rooms have red walls and some have green walls. However, there are *3* rooms in total, so you cannot determine which room you are in by color alone, and two rooms of the same color may have portals that behave differently. You've been into this house many times before. Each time, as you move through the house, you write down what series of portals you take and the color of the room you end up in. You have a collection of paths you've taken where you've ended up in a room with green walls, listed below. Given the final incomplete path at the bottom, write a series of up to 5 remaining steps that will cause you to end up in a room with green walls again.

Tag your final answer like <answer>ab</answer>.

Given these paths that end in a room with green walls:
cbcbabbcca
abcaaacbaa
aabccbabbb
bbbccbbbca
aababaccba
aaaacbacac
baacbccbaa
cbbaacabcc
baabaacaab
bbbbbcacab
acaabcbbba
acaacbccac
cacbabcbba
abcbcbcbcc
ccaccccaba
bcbcabbcca
baabacabca
caababacac
bacacaccaa
bcacbbbbca
bcbbbcaccc
ccabbcccbb
bccbcabbca
baacbabcbc
ccacabccab
caacbcaaab
cacbaaccac
aaccbcaabb
abacabcaab
bacbcbcaca

Complete the following path:
caacb
\end{lstlisting}

\section{Compute Usage} \label{app:compute-cost}

The experiments in this paper on proprietary models had the following (approximate) costs.

\begin{itemize}
    \item gpt-5: \$920
    \item o3-mini: \$430
    \item 4o: somewhere between \$100 and \$200
    \item 4o-mini: somewhere between \$50 and \$150
    \item claude-3.5: \$80
\end{itemize}

The open weight experiments took a cumulative 10-50 GPU-hours on NVIDIA RTX 6000 Ada Generation GPUs, some models requireed the use of 4 in parallel.

\clearpage


\newpage
\section*{NeurIPS Paper Checklist}

\begin{enumerate}

\item {\bf Claims}
    \item[] Question: Do the main claims made in the abstract and introduction accurately reflect the paper's contributions and scope?
    \item[] Answer: \answerYes{}{} 
    \item[] Justification: We claim that the paper contributes a domain that demonstrates that novelty alone can cause LLMs to perform poorly. and in the paper we provide such a domain and comprehensive set of experiments to this effect.

\item {\bf Limitations}
    \item[] Question: Does the paper discuss the limitations of the work performed by the authors?
    \item[] Answer: \answerYes{}{} 
    \item[] Justification: We maintain a fairly narrow scope for the paper in terms of what we are trying to prove. We discuss limitations of our prompt set, which are the main direct limitations.

\item {\bf Theory assumptions and proofs}
    \item[] Question: For each theoretical result, does the paper provide the full set of assumptions and a complete (and correct) proof?
    \item[] Answer: \answerNA{} 
    \item[] Justification: we advance no novel formal theoretical claims in this work.

    \item {\bf Experimental result reproducibility}
    \item[] Question: Does the paper fully disclose all the information needed to reproduce the main experimental results of the paper to the extent that it affects the main claims and/or conclusions of the paper (regardless of whether the code and data are provided or not)?
    \item[] Answer: \answerYes{} 
    \item[] Justification: we believe the results of this paper are as reproducible as is possible. A code repository contains the exact code used to run these experiments, which rely on data generated in that repository. The only significant concern for reproducibility is the availability of certain proprietary models. This issue is unfortunately unavoidable if we are to engage and evaluate proprietary models, so we do not view it as a significant limitation.

\item {\bf Open access to data and code}
    \item[] Question: Does the paper provide open access to the data and code, with sufficient instructions to faithfully reproduce the main experimental results, as described in supplemental material?
    \item[] Answer: \answerYes{} 
    \item[] Justification: A zip file has been attached containing the code.

\item {\bf Experimental setting/details}
    \item[] Question: Does the paper specify all the training and test details (e.g., data splits, hyperparameters, how they were chosen, type of optimizer, etc.) necessary to understand the results?
    \item[] Answer: \answerYes{} 
    \item[] Justification: All information necessary to run the experiments is present in the paper. Many of the details are in Appendix~\ref{app:sampling-of-dfas}.

\item {\bf Experiment statistical significance}
    \item[] Question: Does the paper report error bars suitably and correctly defined or other appropriate information about the statistical significance of the experiments?
    \item[] Answer: \answerYes{}{} 
    \item[] Justification: 95\% CIs are provided everywhere, and our main results (4-Gram through 6-Gram vs everything else) are all significant, as declared in the results section. For more details, see Appedix~\ref{app:significance}.

\item {\bf Experiments compute resources}
    \item[] Question: For each experiment, does the paper provide sufficient information on the computer resources (type of compute workers, memory, time of execution) needed to reproduce the experiments?
    \item[] Answer: \answerYes{} 
    \item[] Justification: See Section~\ref{app:compute-cost}. o1-preview did not go into the paper as it was superceded by o3-mini, which performed better, cost an additional \$180.
    
\item {\bf Code of ethics}
    \item[] Question: Does the research conducted in the paper conform, in every respect, with the NeurIPS Code of Ethics \url{https://neurips.cc/public/EthicsGuidelines}?
    \item[] Answer: \answerYes{} 
    \item[] Justification: This experiment does not intersect any elements of the code of codnudct, except as discussed in the impact statement.

\item {\bf Broader impacts}
    \item[] Question: Does the paper discuss both potential positive societal impacts and negative societal impacts of the work performed?
    \item[] Answer: \answerYes{}{} 
    \item[] Justification: Yes, see post-conclusion impact statement.
    
\item {\bf Safeguards}
    \item[] Question: Does the paper describe safeguards that have been put in place for responsible release of data or models that have a high risk for misuse (e.g., pretrained language models, image generators, or scraped datasets)?
    \item[] Answer: \answerNA{}{} 
    \item[] Justification: No models are being released as part of this paper. The dataset that is being released has no potential for problematic use, and would be fairly easy for someone to create anyway.

\item {\bf Licenses for existing assets}
    \item[] Question: Are the creators or original owners of assets (e.g., code, data, models), used in the paper, properly credited and are the license and terms of use explicitly mentioned and properly respected?
    \item[] Answer: \answerYes{}{} 
    \item[] Justification: We cite every model we use, as well as VLLM. No other artifacts are used.

\item {\bf New assets}
    \item[] Question: Are new assets introduced in the paper well documented and is the documentation provided alongside the assets?
    \item[] Answer: \answerNA{} 
    \item[] Justification: no new assets are released.

\item {\bf Crowdsourcing and research with human subjects}
    \item[] Question: For crowdsourcing experiments and research with human subjects, does the paper include the full text of instructions given to participants and screenshots, if applicable, as well as details about compensation (if any)? 
    \item[] Answer: \answerNA{}{} 
    \item[] Justification: No crowdsourcing or human subjects were included in this paper.

\item {\bf Institutional review board (IRB) approvals or equivalent for research with human subjects}
    \item[] Question: Does the paper describe potential risks incurred by study participants, whether such risks were disclosed to the subjects, and whether Institutional Review Board (IRB) approvals (or an equivalent approval/review based on the requirements of your country or institution) were obtained?
    \item[] Answer: \answerNA{}{} 
    \item[] Justification: No crowdsourcing or human subjects were included in this paper.

\item {\bf Declaration of LLM usage}
    \item[] Question: Does the paper describe the usage of LLMs if it is an important, original, or non-standard component of the core methods in this research? Note that if the LLM is used only for writing, editing, or formatting purposes and does not impact the core methodology, scientific rigorousness, or originality of the research, declaration is not required.
    \item[] Answer: \answerYes{} 
    \item[] Justification: LLM evaluation is the purpose of this paper, and as such LLMs were used throughout the experiments. Their use is precisely described.

\end{enumerate}

\end{document}